\documentclass[nolinenumbers]{iucrjournals}
\usepackage{amsmath,amssymb,amsfonts}
\usepackage{algorithmic}
\usepackage{algorithm}
\usepackage{array}
\usepackage{bm}
\usepackage{textcomp}
\usepackage{stfloats}
\usepackage{url}
\usepackage{verbatim}
\usepackage{graphicx}
\usepackage{textcomp}
\usepackage{xcolor}
\usepackage{dsfont}
\usepackage{bm,mathabx,amsthm}
\usepackage{subcaption}
\usepackage{graphicx}
\usepackage{booktabs}
\usepackage{romannum}

\usepackage{bm}
\usepackage{microtype}
\usepackage{graphicx}
\usepackage{subcaption}
\usepackage{booktabs} 

\usepackage[capitalize,noabbrev]{cleveref}
\usepackage[textsize=tiny]{todonotes}

\usepackage{threeparttable}
\usepackage{wrapfig}
\usepackage{multicol}
\usepackage{multirow}
\usepackage{makecell}

\usepackage{titletoc}
\usepackage[page,header]{appendix}

\title{Beyond Shallow Behavior: Task-Efficient Value-Based Multi-Task Offline MARL via Skill Discovery}

\author{Xun Wang$^{*}$}
\author{Zhuoran Li$^{*}$}
\author{Hai Zhong}
\author{Longbo Huang$^\dag$}

\affil{Institute for Interdisciplinary Information Sciences (IIIS), Tsinghua University \{wang-x24,lizr20,zhongh22\}@mails.tsinghua.edu.cn, longbohuang@tsinghua.edu.cn }

\begin{document} 
\maketitle 
\makeatletter
\renewcommand{\@makefnmark}{}
\makeatother
\footnotetext{$^{*}$ Contributed equally to this work.}
\footnotetext{$^{\dag}$ Corresponding Author.}

\begin{abstract}
  As a data-driven approach, offline MARL learns superior policies solely from offline datasets, ideal for domains rich in historical data but with high interaction costs and risks. However, most existing methods are task-specific, requiring retraining for new tasks, leading to redundancy and inefficiency. To address this issue, we propose a task-efficient value-based multi-task offline MARL algorithm, Skill-Discovery Conservative Q-Learning (SD-CQL). Unlike existing methods decoding actions from skills via behavior cloning, SD-CQL discovers skills in a latent space by reconstructing the next observation, evaluates fixed and variable actions separately, and uses conservative Q-learning with local value calibration to select the optimal action for each skill. It eliminates the need for local-global alignment and enables strong multi-task generalization from limited, small-scale source tasks. Substantial experiments on StarCraft II demonstrate the superior generalization performance and task-efficiency of SD-CQL. It achieves the best performance on $\textbf{13}$ out of $14$ task sets, with up to $\textbf{68.9\%}$ improvement on individual task sets.
\end{abstract}


\section{Introduction}
Multi-agent reinforcement learning (MARL), as a cornerstone of artificial intelligence, provides advanced methodologies to tackle complex challenges requiring coordinated, task-driven decision-making among multiple agents through interaction \cite{shoham2008multiagent,gronauer2022multi}. Integrated with deep neural networks, MARL has demonstrated exceptional success across a diverse range of critical applications, e.g., video games \cite{mathieu2021starcraft}, autonomous systems \cite{shalev2016safe}, and finance \cite{lee2007multiagent}. {As data’s centrality to machine learning grows, }offline MARL has drawn increased attention from researchers \cite{li2023beyond, shao2023counterfactual,liu2024offlinemultiagentreinforcementlearning}. {It learns policies solely from offline datasets, particularly suitable for domains where historical data is abundant but real-time interaction is costly or risky.}

However, most existing offline MARL methods \cite{li2023beyond,shao2023counterfactual,liu2024offlinemultiagentreinforcementlearning} are task-specific. They tend to focus solely on {the source task of the dataset and train a tailored policy. Consequently, although they may achieve impressive performance, any alterations to the deployment environment or the task, such as a small change in the controllable agents count, necessitate reacquiring data and training an entirely new policy, leading to significant inefficiency.}

Hence, unlike existing approaches, we seek to build a method capable of learning versatile policies offline on a small set of known tasks and generalizing well to unseen {but similar} scenarios. By doing so, one can eliminate the redundancy and resource overhead of repeatedly retraining from scratch for each new task. Furthermore, in practice, accessible task scenarios are typically limited, making it infeasible to train policies across all potential scenarios. Hence, we aim for this method to exhibit high \textit{\textbf{task-efficiency}}—handling more unseen tasks {more effectively} with fewer known tasks. {This will help broaden its applicability, especially in scenarios where known tasks are scarce.}

\begin{figure}[!t]
\centering
    \begin{subfigure}{0.47\columnwidth}
        \centering
        \includegraphics[width=\textwidth]{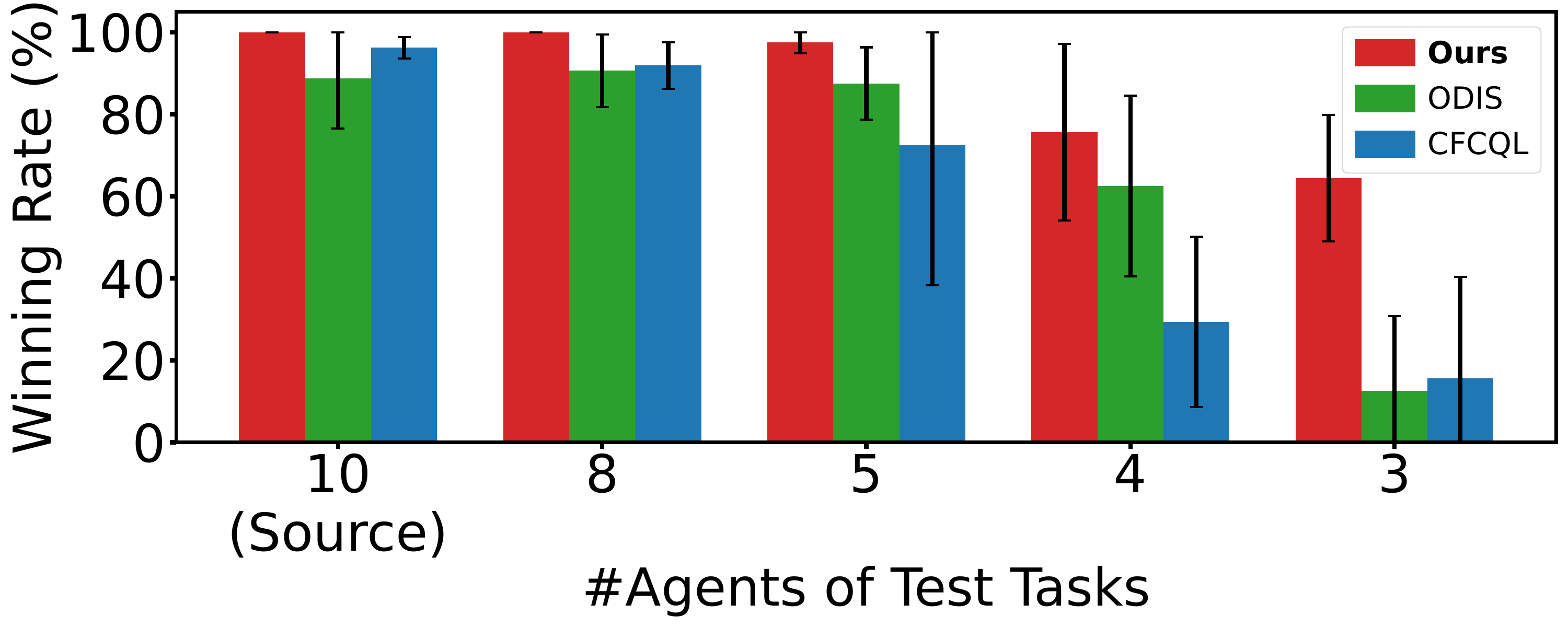}
        \caption{Skill-Discovery enhances the multi-task generalization. {Trained on \textit{10m} dataset, and tested on all tasks.} }
        \label{fig:inv}
    \end{subfigure}
    \hspace{0.4cm}
    \begin{subfigure}{0.47\columnwidth}
        \centering
        \includegraphics[width=\textwidth]{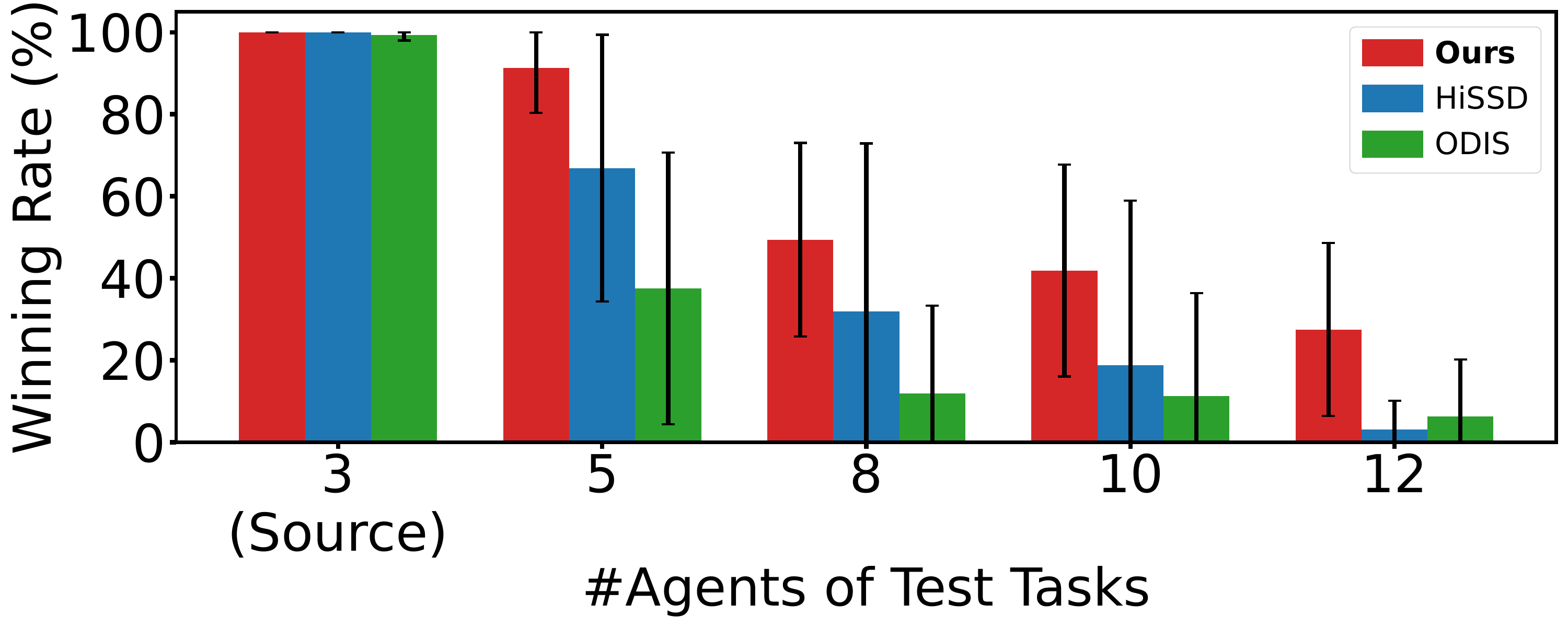}
        \caption{Recent Skill-Discovery methods struggle due to the limited source task. {Trained on \textit{3m} dataset, and tested on all tasks.}}
        \label{fig:single}
    \end{subfigure}
\caption{Skill-Discovery enhances cross-task generalization for offline MARL, while existing methods have limitations. The number with \textit{Source} behind it indicates the source task used for training. {The error bars represent $\pm 1$ standard deviation (truncated outside the $0\sim100\%$ range).}}
\label{fig:moti}

\end{figure}

Therefore, we need to confront the following two key challenges: {(i) The distributional shift between limited offline data and the real environmental dynamics within each task. (ii) Generalization across tasks, that is, maintaining policy performance on unseen tasks without access to their offline data or any dynamic information.} Although we can overcome the first challenge via conservative policy optimization, increased conservatism on the source task usually leads to poorer generalization to unseen tasks, highlighting the importance of solving both challenges simultaneously. 

As shown in Figure \ref{fig:inv}, we train the advanced offline MARL algorithm \cite{shao2023counterfactual} on a task with $10$ agents (source task) and test it on tasks with fewer agents. We find that it performs poorly when the number of agents in the test tasks is reduced, even though these tasks appear easier. This suggests that the agents need to learn and utilize generalizable decision-making structures across tasks, making skill-discovery a promising solution. In this approach, agents identify high-level decision patterns, i.e., ``skills'', from source tasks and select the appropriate one to perform specific actions in unseen tasks, achieving cross-task generalization. However, most existing {skill-discovery} methods {require online interactions to} discover skills \cite{hsl2022,tian2024decompose}{, or learn the utilization of offline discovered skills \cite{chen2024variational}}, with limited attention to the {entirely} offline setting. 

Recently, ODIS \cite{zhang2023odis} {and HiSSD \cite{hissd} propose methods} for multi-task offline MARL, but as shown in Figure \ref{fig:single}, {they both struggle} to discover effective skills in small-scale source tasks. {Moreover, both methods rely on behavior cloning (BC) to decode actions from a presumed optimal skill representation directly. However, as a skill usually spans a subset of offline actions, BC biases agents toward likely, not optimal, choices. While optimizing actions via} {value-based methods} {can mitigate this likelihood bias, challenges in the scalability and generalization of Q-values leave its feasibility for offline multi-task MARL still to be established.} This motivates us to investigate the following important question: 

\textbf{\textit{Can we {combine the strengths of skill discovery and} {value-based methods} {to enable} strong cross-task performance with high task-efficiency?}} 

\begin{figure}[!t]
\centering
    \begin{subfigure}[b]{0.3\columnwidth}
        \centering
        \includegraphics[width=.9\textwidth]{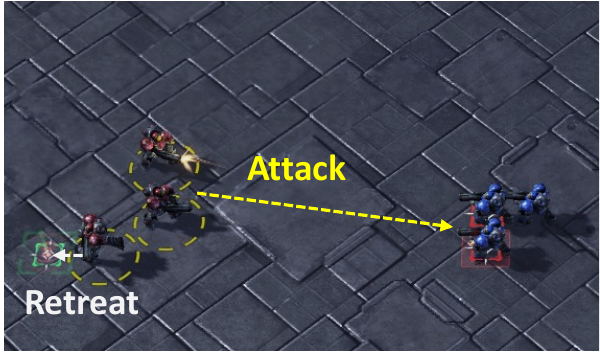}
        \caption{Training in \textit{3m}.}\label{fig:3m_visual}
    \end{subfigure}
    \begin{subfigure}[b]{0.3\columnwidth}
        \centering
        \includegraphics[width=.9\textwidth]{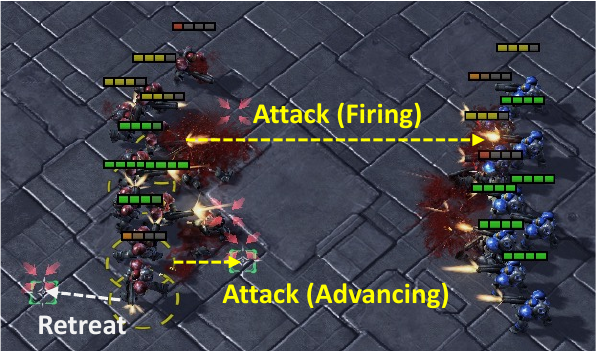}
        \vspace{-.5mm}
        \caption{Testing in \textit{12m}.}\label{fig:12m_visual}
    \end{subfigure}
    \begin{subfigure}[b]{0.3\columnwidth}
        \centering
        \includegraphics[height=0.6\columnwidth]{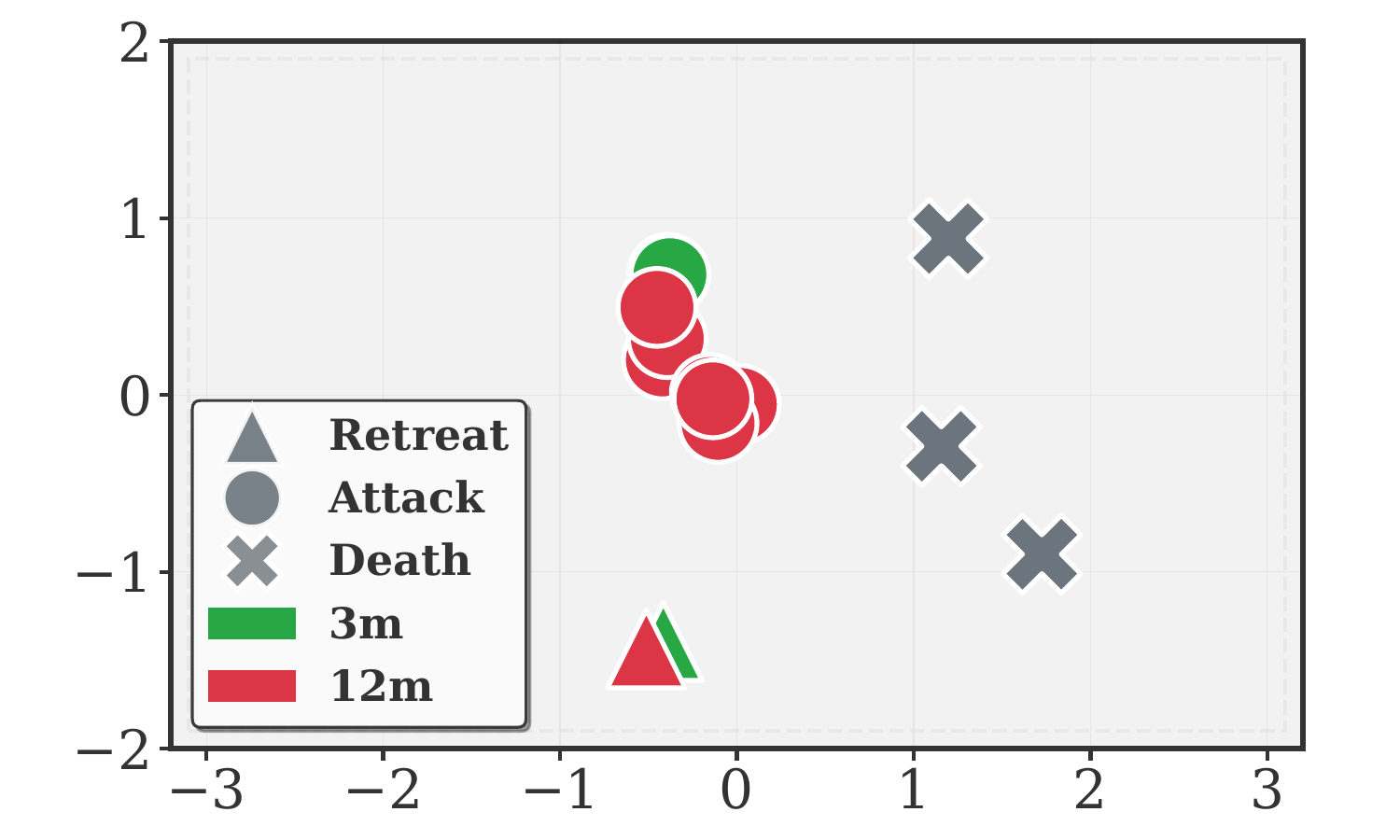}
        \vspace{-.6cm}
        \caption{Visualization of skills.}\label{fig:z_visual}
    \end{subfigure}
    \caption{SD-CQL effectively learns two skills, \textbf{\textit{retreat}} and \textbf{\textit{attack}}, from the source task (a) and successfully transfers them to the target task (b), each with distinct actions. This is consistent with the visualization in (c), where each marker corresponds to a skill adopted by the agent.}
\label{fig:visual}
\end{figure}

We answer this question affirmatively by proposing the \underline{S}kill-\underline{D}iscovery \underline{C}onservative \underline{Q}-\underline{L}earning algorithm (\textbf{SD-CQL}). It achieves effective skill discovery by reconstructing the {next local observation} and training a set of {scalable} skill-conditioned policies {via local value calibration}{, showing that} {value-based method} {in this setting is not only viable but also superior.} {As} shown in Figure \ref{fig:visual}, the offline agents acquire two skills, \textit{retreat} and \textit{attack}, from the 3-marines task (Figure \ref{fig:3m_visual}). When deployed in the 12-marines task (Figure \ref{fig:12m_visual}), they exhibit the ability to \textit{retreat} when health is low and to \textit{attack} when health is sufficient, with \textit{attack} encompassing two specific actions: firing and advancing. It is consistent with the visualization of the skills adopted by the agent, as shown in Figure \ref{fig:z_visual}. More details can be found in Section \ref{exp_visual} and Appendix \ref{app:visual}.

Different from existing methods \cite{zhang2023odis, hissd} that rely on {global information} or predefine {the skill count, SD-CQL discovers continuous skill vectors entirely based on local observations.} {This prompts agents to extract expressive, task-agnostic features from limited data, and eliminates the local-global alignment. Then, SD-CQL conservatively trains separate skill-conditioned Q-networks for fixed and variable actions, and applies local value calibration to mitigate agent-scaling estimation error. Extensive experiments demonstrate its SOTA task efficiency.}

{This paper is organized as follows. Section \ref{sec:bg} introduces the necessary background, followed by Section \ref{sec:method}, which elaborates on the proposed algorithm. Section \ref{sec:exp} presents the main empirical results, and Section \ref{sec:con} concludes the paper with key takeaways. Beyond the main text, the appendices provide a comprehensive review of related work (Appendix \ref{app:rw}), implementation and experimental details (Appendices \ref{app:algo} and \ref{app:exp_detail}), additional results (Appendices \ref{app:visual}, \ref{app:ablation}, \ref{app:results}, and \ref{app:curve}), {and} extended discussions (Appendix \ref{app:dis}).}

{In summary, the key contributions of this work are:}
\begin{itemize}
    \setlength{\itemsep}{0pt}
    \item We investigate a rarely explored {yet important} field: multi-task offline MARL. {We analyze current limitations and propose a novel method to demonstrate that} {value-based method} {is not only viable but also superior in this setting}, offering a new perspective for research in this area.

    \item We propose SD-CQL for multi-task offline MARL. {It} discovers skills {in latent space} by reconstructing the next observation{, then} evaluates fixed and variable actions separately. {Finally, it executes the optimal action for each skill via conservative Q-learning with local value calibration, enabling better generalization and task efficiency than existing methods.} 
    \item Through comprehensive experiments, we show that SD-CQL achieves the best performance in {$\textbf{13}$} out of $14$ task sets, with an average performance improvement of up to {$\textbf{68.9\%}$} on individual task sets over existing baselines.
\end{itemize}
\section{Background}\label{sec:bg}
\subsection{Multi-task MARL and Multi-Task Offline MARL} A cooperative MARL task, indexed by \( m \), can be formulated as a decentralized partially observable Markov decision process (Dec-POMDP) \cite{oliehoek2016concise}. A Dec-POMDP is represented by a tuple \( (\mathcal{N}_{m}, \mathcal{S}_{m}, \mathcal{O}_{m}, \mathcal{A}_{m}, P_{m}, R_{m}, \gamma) \), where \( \mathcal{N}_{m} \) is the set of agents, \( \mathcal{S}_{m} \) is the set of states, \( \mathcal{O}_{m} \) is the joint observation space, \( \mathcal{A}_{m} \) is the joint action space, \( P_{m} \) is the state transition probability (defining the probability of transitioning to the next state given the current state and joint action), \( R_{m} \) is the immediate reward shared by all agents, and \( \gamma \) is the discount factor.

The multi-task MARL problem is a collection of such MARL tasks, which can be represented by the tuple \( (\mathcal{N}, \mathcal{S}, \mathcal{O}, \mathcal{A}, P, R, \gamma, \mathbb{T}) \). Here, \( \mathbb{T} \) is the set of tasks, \( \gamma \) is the discount factor shared by all tasks, and the remaining elements are the union of corresponding elements across all tasks. For example, \( \mathcal{N} = \bigcup_{m \in \mathbb{T}} \mathcal{N}_m \) represents the union of the sets of agents across all tasks.

For the task $m$, each agent maintains its observation-action history \( \tau^{i} \in \mathcal{T}^{i}_{m} \), and the corresponding joint history is denoted by \( \tau \in \mathcal{T}_{m} \). \( \mathcal{T} = \bigcup_{m \in \mathbb{T}} \mathcal{T}_{m} \) represents the collection of observation-action histories across all tasks, and the goal is to find a joint policy \( \pi: \mathcal{T}\to \mathcal{A} \) that maximizes the expected discounted return average over all tasks:
\(
\mathcal{J}=\mathbb{E}_{\pi,\mathbb{T}} \left[ \sum_{t=0}^{T} \gamma^t \cdot r_m(s^m_t, \boldsymbol{a}^m_t) \right]
\),
where $T$ is the time horizon and $r_m(s^m_t, \boldsymbol{a}^m_t)$ is the reward for taking joint action $\boldsymbol{a}^m_t$ at state $s^m_t$ in task $m$.

As for multi-task offline MARL, agents can only access the decision dataset \( \mathcal{D}=\{\mathcal{D}_m\}_{m\in\mathbb{T}_s} \), where \( \mathbb{T}_s \subset \mathbb{T} \) is referred to as \textit{source tasks} (the complement referred to as \textit{unseen tasks}), without any interaction with the environment. The goal is still to maximize $\mathcal{J}$. 
In Appendix \ref{app:multi}, {we further elaborate on the multi-task learning in the contexts of MARL and conventional RL.} 

\subsection{CTDE Framework{, QMIX and CQL}}

A common framework for cooperative MARL is Centralized Training for Decentralized Execution (CTDE). In this framework, agents can leverage centralized information during training but must rely solely on their local observations during execution. In this work, we adopt QMIX \cite{qmix} as our backbone algorithm, which is one of the most popular discrete-action CTDE algorithms. In principle, our method can be applied to any value-based algorithm.

In QMIX, each agent maintains an individual Q-function \( Q_i(\tau^i, a^i) \), which is conditioned on its own observation-action history \( \tau^i \) and action \( a^i \). Then, it calculates a joint Q-function \( Q_{tot}(\tau, \boldsymbol{a}) \) from individual Q-functions through a mixing network \( f_s \), such that:
\begin{equation}\label{eq:qmix}
Q_{\text{tot}}(\tau, \boldsymbol{a}) = f_s(Q_1(\tau^1, a^1), \ldots, Q_n(\tau^n, a^n)).
\end{equation}
The mixing network is designed to satisfy the monotonicity constraint, ensuring that the partial derivative of \( Q_{\text{tot}} \) with respect to each \( Q_i \) is non-negative. To train the Q-function, QMIX minimizes the temporal-difference error on \( Q_{\text{tot}} \).

{In offline MARL, the lack of interaction may cause Q-value overestimation on out-of-distribution (OOD) samples. Thus, Conservative Q-learning (CQL) \cite{cql} uses a regularization term to enhance the Q-value of in-dataset samples while suppressing that of OOD ones:}
\begin{equation}
{\mathcal{L}_{\text{CQL}}=\mathbb{E}_{\tau \sim \mathcal{D}, \boldsymbol{a} \sim \mu} \left[ Q_{tot}(\tau, \boldsymbol{a}) \right] - \mathbb{E}_{(\tau, \boldsymbol{a}) \sim \mathcal{D}} \left[ Q_{tot}(\tau, \boldsymbol{a}) \right]} \label{CQL}
\end{equation}
{where $\mu$ denotes the sampling distribution (e.g., sampling from a uniform distribution).}

\section{Skill-Discovery Conservative Q-Learning}\label{sec:method}

\begin{figure*}[t!]
\centering
\centerline{\includegraphics[width=1.0\textwidth]{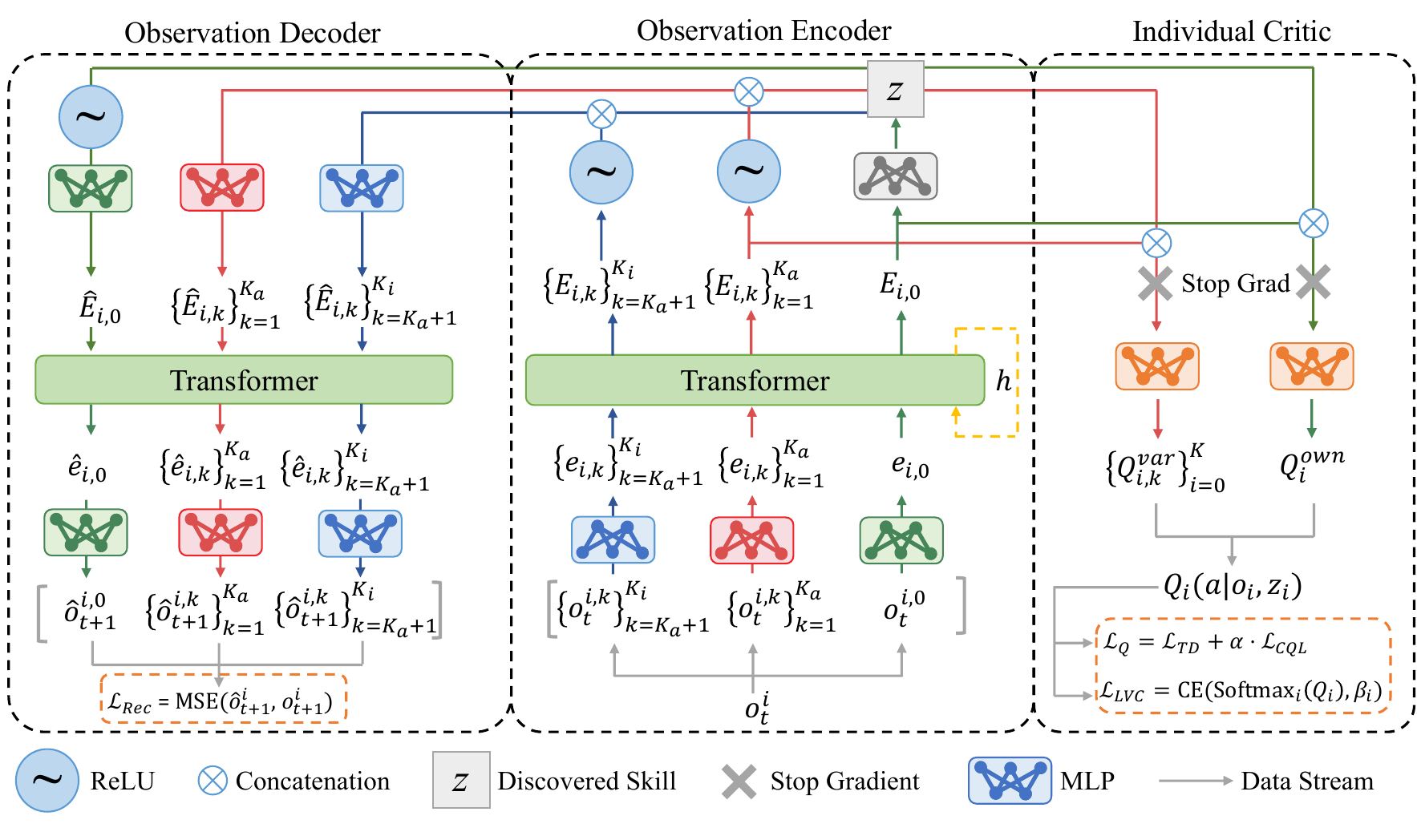}}
\caption{SD-CQL framework. For the $i$-th agent, the encoder splits the current observation \( o_t \) into \( K_i + 1 \) entities, embedding them into consistent dimensions, with the \(0\)-th representing the agent itself. Then, the skill $z$ is extracted from $E_{i,0}$, the embedding associated with the agent itself. Finally, SD-CQL injects local information into \(z\) by reconstructing the next observation \( o_{t+1} \). Meanwhile, it computes the \(z\)-conditioned Q-values for fixed and variable actions separately, and optimizes scalable polices via CQL and {local value calibration}.
} 
\label{fig:sdcql}
\vskip -0.2in
\end{figure*}

To develop scalable policies that can generalize to varying unseen tasks through data from only a few source tasks, we propose Skill-Discovery Conservative Q-Learning (SD-CQL), a task-efficient algorithm for multi-task offline MARL. As illustrated in Figure \ref{fig:sdcql}, SD-CQL discovers cross-task generalizable skills in the latent space through observation reconstruction, and then trains scalable policies via multi-task CQL and {local value calibration}. This approach mitigates distributional shifts and error accumulation in offline MARL, as well as the generalization challenges inherent in multi-task learning. 
{We present a detailed introduction of each component below, and offer an in-depth architectural comparison with existing methods in Appendix \ref{app:comp}.} The full algorithm is presented in Algorithm \ref{alg:sdcql}{, located in Appendix \ref{app:algo}.}

\subsection{Observation Reconstruction}
To improve policy generalization across tasks, a natural idea is to make agents learn task-shared decision features, i.e., ``skills'', from offline data. Hence, we choose to achieve this by reconstructing the next {local} observations without global state or task-specific rewards. {This not only helps agents to learn more transferable, task-agnostic temporal features, but also reduces skills’ dependence on task scale and enables direct distributed execution without local-global alignment.} 
\subsubsection{Observation Encoder}
As the number of agents may vary across tasks, leading to changes in state or observation size, {Following the standard practice \cite{pmlr-v100-liu20a, zhang2023odis, wu2024portal}, we decompose the state or observation into entity units and map them into a shared embedding space.} We then adopt a transformer \cite{vaswani2017attention}, following \cite{zhang2023odis}, for further processing.

Specifically, we first decompose the $i$-th agent's observation at $t$-th time step $o^i_t$ into {$o_t^{i,0}$, which is related to the agent itself, and $\{o_t^{i,k}\}_{k=1}^{K_i}$ corresponding to the entities it can observe, among which $K_a$ entities are interactable.} We further encode these heterogeneous entity information into embeddings {$\{e_{i,k}\}_{k=0}^{K_i}$} with same dimensions. Finally, we use a single-layer transformer to capture the relationships between entities and obtain the final encoding vector $\{E_{i,k}\}_{k=0}^{K_i}$ for each entity. {More details of the decomposer and architecture are provided in Appendix \ref{app:imp_detail}.}

{Since} the agents' local observations are relatively limited, we {add an additional entry {$h=e_{i, K_i+1}$} as hidden variable}. {It is recurrently involved in the encoder transformer but is excluded from the decoder, enabling the skill to capture temporal information.} We also incorporate the previous actions \( a_{t-1} \) in \( o_t \), but exclude them from the reconstruction target \( o_{t+1} \) {in line with the standard practices.}

\subsubsection{Observation Decoder}
{To enable situation-appropriate skill selection, we propose a decoder to learn expressive representations in a continuous latent space via reconstructing the next local observation. By focusing on action-agnostic local transitions, these representations capture higher-level spatiotemporal patterns while reducing task-specific information, thereby improving cross-task generalization.} 

Concretely, for the embeddings $\{E_{i,k}\}_{k=0}^{K_i}$ from the observation encoder, {the decoder first extracts a $d$-dimension latent skill vector \( z \in \mathbb{R}^d\). Then, it reconstructs the embeddings $\{\widehat{E}_{i,k}\}_{k=0}^{K_i}$ with $z$. }Finally, similar to the encoder, the reconstructed embeddings are passed through a single-layer transformer, restored into entity vectors \(\{\hat{o}_{i,k}\}_{k=0}^{K_i}\), and reassembled into the next observation \( \hat{o}^i_{t+1} \). {Detailed architecture can be checked in Appendix \ref{app:imp_detail}. Then}, the final reconstruction loss is:
\begin{equation}\label{rec_loss}
\mathcal{L}_{\text{Rec}}=\dfrac{1}{N}\sum_{i=1}^{N}\text{MSE}(\hat{o}_{t+1}^i,o_{t+1}^i)
\end{equation}
where $N$ is the number of {controllable} agents in the task, and $\text{MSE}(\cdot,\cdot)$ {is} the Mean Square Error. 
\subsection{Skill-conditioned Policy Optimization}
After identifying and selecting the appropriate skill, existing multi-task offline MARL algorithms primarily execute the corresponding actions through behavior cloning (BC). However, since a single skill often encompasses multiple specific actions, selecting the optimal skill does not necessarily ensure that all associated actions are optimal.

Therefore, we leverage conservative Q-learning to optimize the skill-conditioned policy $\pi(a\mid s,~z)$, enabling the execution of optimal actions associated with the selected skill. To mitigate the accumulated estimation errors inherent in Q-learning and reduce distributional shift, SD-CQL separately evaluates fixed and variable actions with {local value calibration}, thereby enhancing training stability for larger-scale tasks. 

\subsubsection{{Conservative} Skill-conditioned Q-value}
To handle the variation in the number of actions across different tasks, we utilize two separate Q-networks. One network, called $Q^{own}$, is responsible for a fixed set of actions related to the agent itself, while the other, called $Q^{var}$, handles a variable set of actions associated with other entities in its observation. Both networks receive the corresponding entity embeddings $E_{i,k}$ and skill vectors $z$ as inputs and output the Q-values for their respective actions. The parameters of Q-value networks are shared by all the agents. Therefore, the individual Q-values for the $i$-th agent with skill $z_i$ are: 
\begin{equation}
Q_i(a\mid o_i, z_i)=\begin{cases}
Q^{own}(a\mid E_{i,0},~z_i) &\text{if }a \in \mathcal{A}_i^{own} \\ 
Q^{var}(a\mid E_{i,k},~z_i) &\text{if }a \in \mathcal{A}_{i}^{k} 
\end{cases}
\end{equation}
where $\mathcal{A}_i^{own}$ is the set of actions that only related to the $i$-th agent itself, and  $\mathcal{A}_{i}^{k}$ is the set of actions that related to the $i$-th agent and the $k$-th other entity. {While similar techniques have been adopted in previous work \cite{lin2024hgap} to enhance Q-value estimation, it is important to note that our use of separate networks is primarily motivated by the need to accommodate the varying dimensions of states, observations, and actions inherent in multi-task offline MARL.}

To avoid interference between observation reconstruction and policy optimization, we truncate the gradient propagation of the embeddings and skill vectors before feeding them into the Q-network.

Finally, we employ a mixing network to aggregate the individual Q-values into a global Q-value $Q_{\text{tot}}$ according to \eqref{eq:qmix}. Then, we can optimize $Q_{\text{tot}}$ through the temporal difference loss outlined below:
\begin{equation}\label{td_loss}
\mathcal{L}_{\text{TD}}=\left[ Q_{\text{tot}}({\tau}, \boldsymbol{a} \mid \boldsymbol{z}) - \left( r + \gamma \max_{\boldsymbol{a}'} \bar{Q}_{\text{tot}}({\tau}', \boldsymbol{a}' \mid \boldsymbol{z}') \right) \right]^2
\end{equation}
where \( {\tau}' \), \( \boldsymbol{a}' \) and \( \boldsymbol{z}' \) denote the next observation-action history, joint action, and joint skill, respectively, and $\bar{Q}_{\text{tot}}$ represents the target joint action-value function.

{In the offline setting, the scarcity of data and the lack of online interaction may cause agents to make detrimental decisions by overestimating OOD state–action pairs. To address this issue, we employ CQL \cite{cql} to regularize the skill-conditioned Q-values:}
\begin{equation}
\mathcal{L}_{\text{CQL}}=\mathbb{E}_{\tau \sim \mathcal{D}, \boldsymbol{a} \sim \mu} \left[ Q(\tau, \boldsymbol{a}\mid \boldsymbol{z}) \right] - \mathbb{E}_{(\tau, \boldsymbol{a}) \sim \mathcal{D}} \left[ Q(\tau, \boldsymbol{a}\mid \boldsymbol{z}) \right] \label{CQL_Regularization}
\end{equation}
where $\mu$ denotes the sampling distribution (e.g., sampling from a uniform distribution).
And the total loss function for Q-learning is:
\begin{equation}\label{q_learning}
\mathcal{L}_{\text{Q}} = \mathcal{L}_{\text{TD}} + \alpha \mathcal{L}_{\text{CQL}}
\end{equation}
where $\alpha > 0$ is a hyperparameter to control conservatism.
\subsubsection{{Local Value Calibration}}
In addition to distributional shifts, the accumulation of estimation errors caused by multiple agents is another challenging issue. In multi-agent environments, relying solely on Q-learning remains inadequate to mitigate the rapidly escalating Q-value estimation errors \cite{res}. 

Therefore, inspired by previous studies \cite{td3bc, omar}, {we propose \textit{Local Value Calibration (LVC)} to calibrate each agent’s local Q-value distribution over all available actions against its corresponding behavior policy. This term enhances the stability of SD-CQL across tasks with different scales}:
\begin{equation}\label{bc-term}
\mathcal{L}_{\text{LVC}} = \dfrac{1}{N}\sum_{i=1}^{N}\text{CE}\left(\text{Softmax}(Q_i),~\beta_i\right)
\end{equation}
where $\text{CE}\left(\cdot,\cdot\right)$ is the cross entropy loss, $N$ is the number of agents, $Q_i$ is the individual Q-values of all the actions available for the $i$-th agent, and $\beta_i$ is the corresponding one-hot action in the offline datasets.

Then, the total loss function for SD-CQL is:
\begin{equation}\label{total_loss}
\mathcal{L} = (1-\eta)\cdot\mathcal{L}_{\text{Q}} + \eta \cdot \mathcal{L}_{{LVC}} + \mathcal{L}_{\text{Rec}}
\end{equation}
where $\eta \in [0,1]$ is the hyperparameter to control the strength of {value calibration}.

\section{Experiments}\label{sec:exp}
To evaluate the performance of SD-CQL in multi-task offline MARL scenarios, we establish multiple transfer training task sets based on the StarCraft Multi-Agent Challenge (SMAC) \cite{samvelyan19smac}. Through these experimental task sets, we aim to address the following questions: (i) Compared to existing algorithms, does SD-CQL demonstrate better performance in multi-task offline MARL (Section \ref{exp_eval}), (ii) Do skill vectors indeed characterize the decision-making features of agents in different contexts (Section \ref{exp_visual}), and (iii) Are components within SD-CQL critical to its performance (Appendix \ref{app:ablation}). 
\subsection{Setup}

\paragraph{Datasets} Following the definition by D4RL \cite{fu2020d4rl}, our experiments utilize datasets of four quality: \textit{Expert}, \textit{Medium}, \textit{Medium-Replay}, and \textit{Medium-Expert}. For fair comparisons, we use the datasets provided by ODIS \cite{zhang2023odis}. Details are provided in Appendix \ref{app:datasets}.

\paragraph{Task Sets} 
To comprehensively simulate different transfer scenarios, {we construct 14 representative offline-training, zero-shot transfer task sets across five scenarios: }
\textit{Marine-Easy}, \textit{Marine-Hard}, \textit{Stalker-Zealot}, \textit{Marine-Single}, and \textit{Marine-Single-Inv}. They can be categorized into two groups:  (i) Training on multiple tasks and test on multiple tasks (Multi-to-Multi) and (ii) Training on one task and test on multiple tasks (One-to-Multi). The first three {scenarios}, which fall under the Multi-to-Multi {group}, are consistent with those in ODIS \cite{zhang2023odis}, while the last two, classified as One-to-Multi, are additionally designed by us. More details are included in Appendix \ref{app:tasks}.

\paragraph{Baselines}
We primarily select four offline multi-task MARL algorithms as baselines: BC-t, BC-r, {CQL} \cite{cql}, {ODIS \cite{zhang2023odis}, and HiSSD \cite{hissd}}. Among them, BC-t is a behavior cloning approach based on a multi-task transformer, BC-r incorporates return-to-go information into the input of BC-t, {CQL is a representative offline} {value-based} {method, which we implement with a multi-task transformer,} ODIS is the {first algorithm tailored for multi-task offline MARL, and HiSSD is the state-of-the-art algorithm for multi-task offline MARL. Implementation details for the baselines are provided in Appendix \ref{app:imp_detail}.}

\paragraph{Experiment Setup} 
We implement SD-CQL based on the PYMARL2 \cite{pymarl2} and the ODIS codebase, while the {baselines} are directly utilized from the codes implemented by ODIS \cite{zhang2023odis} {and HiSSD \cite{hissd}}. To ensure fairness, each algorithm runs $35,000$ steps of multi-task offline training. The exception is ODIS, which additionally requires an initial pre-training of $15,000$ steps. During evaluation, each test environment runs $32$ episodes, and the average results are recorded. Unless otherwise specified, we report the average performance of the final policy across five random seeds, with the best performance for each task highlighted in bold. The learning curves are presented in Appendix \ref{app:curve}.

\paragraph{{Hyperparameters}} For SD-CQL, the primary hyperparameter adjusted is the {LVC} weight $\eta$, {while} the CQL weight $\alpha$ {is} fixed at $1.0$ {(the same value as in CQL)}.  As to ODIS {and HiSSD}, we use the hyperparameter configurations provided in {their original papers} \cite{zhang2023odis,hissd} for Multi-to-Multi task sets. For One-to-Multi task sets we specifically designed, {they do} not offer hyperparameter configurations. Thus, we adjust {the primary hyperparameters involved in their RL losses and report the best performance achieved.} All other hyperparameters remain consistent with the default settings. For implementation details and main hyperparameter settings, please refer to Appendix \ref{app:hyperparam}.

\begin{table*}[!ht]
\caption{Win rates of Multi-to-Multi task sets. The reported results are the average performance over all source tasks and unseen tasks, and are averaged over 5 random seeds. {Bold numbers indicate results within 2\% of the best performance,} and $\pm$ denotes one standard deviation.}\label{tab:mm-win-rates}
\vskip 0.15in
\centering
\begin{small}
\begin{tabular}{lccccc}
\toprule
\textbf{Task Set} & {\textbf{BC-best}$^*$}&{\textbf{CQL}}& \textbf{ODIS} & \textbf{HiSSD} & \textbf{SD-CQL (Ours)} \\
\midrule
\textit{marine-easy-e} & \textbf{99.38} $\pm$ 0.40  &31.13 $\pm$ 6.36& 70.50 $\pm$ 30.14 & \textbf{97.94} $\pm$ 1.23 & \textbf{98.06} $\pm$ 1.00   \\
\textit{marine-hard-e} & 64.84 $\pm$ 3.24  &24.84 $\pm$ 1.53& 21.09 $\pm$ 14.72 & 68.02 $\pm$ 4.47 & \textbf{70.68} $\pm$ 4.80 \\
\textit{stalker-zealot-e} & 68.61 $\pm$ 3.93 &10.62 $\pm$ 5.88& 55.05 $\pm$ 9.11 & 66.59 $\pm$ 7.77 & \textbf{75.72} $\pm$ 3.73\\
\midrule
\textit{marine-easy-m} & 71.81 $\pm$ 3.74 &17.38 $\pm$ 7.31& 68.50 $\pm$ 5.74 & 71.44 $\pm$ 2.77 & \textbf{75.81} $\pm$ 1.73\\
\textit{marine-hard-m} & 42.08 $\pm$ 2.63 &9.17 $\pm$ 1.78& 32.55 $\pm$ 4.31 & \textbf{47.60} $\pm$ 2.66 & \textbf{48.12} $\pm$ 5.66 \\
\textit{stalker-zealot-m} & 23.94 $\pm$ 2.72 &15.29 $\pm$ 4.89& 11.97 $\pm$ 7.53 & 23.51 $\pm$ 3.76 & \textbf{40.43} $\pm$ 7.06 \\
\midrule
\textit{marine-easy-mr} & 49.62 $\pm$ 9.17  &29.56 $\pm$ 11.73& 11.06 $\pm$ 8.94 & 72.88 $\pm$ 3.16 & \textbf{75.81} $\pm$ 6.83 \\
\textit{marine-hard-mr}  & 46.98 $\pm$ 1.95  &21.67 $\pm$ 2.83& 37.29 $\pm$ 5.48 & 47.55 $\pm$ 3.56 & \textbf{49.84} $\pm$ 3.36\\
\textit{stalker-zealot-mr} & 17.64 $\pm$ 4.40  &14.57 $\pm$ 2.02& 8.17 $\pm$ 6.08 & 15.00 $\pm$ 4.02 &\textbf{23.08} $\pm$ 3.38 \\
\midrule
\textit{marine-easy-me} & 75.25 $\pm$ 7.41  &33.56 $\pm$ 6.47& 66.00 $\pm$ 9.21 & 78.31 $\pm$ 4.99 & \textbf{86.88} $\pm$ 5.68\\
\textit{marine-hard-me} & 51.93 $\pm$ 7.21 &20.31 $\pm$ 1.97& 19.58 $\pm$ 17.63 & 53.28 $\pm$ 1.88 & \textbf{56.20} $\pm$ 3.21 \\
\textit{stalker-zealot-me} & 39.62 $\pm$ 2.98  &13.41 $\pm$ 2.53& 30.72 $\pm$ 14.65 & 40.29 $\pm$ 4.64 & \textbf{66.97} $\pm$ 7.01  \\
\bottomrule
\multicolumn{6}{l}{\small \textit{-e}, \textit{-m}, \textit{-mr}, and \textit{-mr} represent \textit{-expert}, \textit{-medium}, \textit{-medium-replay}, and \textit{-medium-expert}, respectively.}\\
\multicolumn{6}{l}{{\small $^*$ BC-best represents the better average performance between BC-r and BC-t.}}
\end{tabular}
\end{small}
\vskip -0.1in
\end{table*}

\subsection{Evaluation}\label{exp_eval}

\subsubsection{Multi-to-Multi}

The Multi-to-Multi task sets include \textit{Marine-Easy}, \textit{Marine-Hard}, and \textit{Stalker-Zealot}, each comprising three source tasks of varying scales and several unseen tasks. Agents are trained offline using datasets from the source tasks and are tested across all tasks, with zero-shot testing conducted on the unseen tasks. For each task set, we set up four datasets of different qualities. We report the average performance of the algorithms across all tasks for each task set and dataset in Table \ref{tab:mm-win-rates}{, where BC-best denotes the better result between BC-t and BC-r}. Detailed results can be found in Appendix \ref{app:results-mtm}.

The results show that SD-CQL achieves the best performance for {$11$} out of $12$ task sets, with the {remaining one being very close to the best baseline algorithms (within $1.2\%$).} In particular, {in \textit{Stalker-Zealot-Medium} and \textit{Stalker-Zealot-Medium-Expert} task sets, SD-CQL significantly outperforms other algorithms, with performance improvements of $68.9\%$ and $66.2\%$}, respectively, compared to the best-performing baseline. This demonstrates that SD-CQL is capable of maintaining superior multi-task offline generalization performance across various dataset qualities. 

\subsubsection{One-to-Multi}

\begin{table*}[!ht]
\caption{Win rates of One-to-Multi task sets. The results are averaged over 5 random seeds. {Bold numbers indicate results within 2\% of the best performance,} and $\pm$ denotes one standard deviation.}\label{tab:om-win-rates}
\centering
\resizebox{\linewidth}{!}{%
\begin{small}
\begin{tabular}{clcccccc}
\toprule
\multicolumn{2}{c}{\textbf{Task Set}} & {\textbf{BC-best$^*$}} & {\textbf{CQL}} & \textbf{ODIS} & \textbf{HiSSD} & \textbf{SD-CQL (Ours)} \\
\midrule
\multirow{5}{*}{\rotatebox{90}{\makecell{\textit{Marine}\\\textit{Single}}}} 
    &3m $^\diamond$ & \textbf{100.00} $\pm$ 0.00 & \textbf{100.00} $\pm$ 0.00 & \textbf{99.38} $\pm$ 1.4 & \textbf{100.00} $\pm$ 0.00 & \textbf{100.00} $\pm$ 0.00\\
    &5m & 81.88 $\pm$ 9.22 & 88.12 $\pm$ 7.46 & 37.5 $\pm$ 33.15 & 66.88 $\pm$ 32.52 & \textbf{91.25} $\pm$ 10.92\\
    &8m & 38.75 $\pm$ 22.38 & 40.00 $\pm$ 26.00 & 11.88 $\pm$ 21.47 & 31.87 $\pm$ 41.01 & \textbf{49.38} $\pm$ 23.63\\
    &10m & 20.62 $\pm$ 22.38 & 34.38 $\pm$ 25.39  & 11.25 $\pm$ 25.16 & 18.75 $\pm$ 40.20 & \textbf{41.88} $\pm$ 25.83\\
    &12m & 9.38 $\pm$ 13.98 & 17.50 $\pm$ 10.27 & 6.25 $\pm$ 13.98 & 3.12 $\pm$ 6.99 & \textbf{27.50} $\pm$ 21.13\\
    \midrule
    \multicolumn{2}{c}{\textbf{Average}} & 50.12 $\pm$ 9.65 & 56.00 $\pm$ 11.07 & 33.25 $\pm$ 16.50 & 44.12 $\pm$ 18.89 & \textbf{62.00} $\pm$ 13.80\\
\midrule\midrule
\multirow{5}{*}{\rotatebox{90}{\makecell{\textit{Marine} \\ \textit{Single-Inv}}}}
    &10m $^\diamond$ & \textbf{100.00} $\pm$ 0.00 & 0.62 $\pm$ 1.4 &  88.75 $\pm$ 12.22 & \textbf{100.00} $\pm$ 0.00 & \textbf{100.00} $\pm$ 0.00\\
    &8m &  98.75 $\pm$ 1.71 & 3.75 $\pm$ 8.39 & 90.62 $\pm$ 8.84 & \textbf{100.00} $\pm$ 0.00 & \textbf{100.00} $\pm$ 0.00\\
    &5m & 33.12 $\pm$ 33.63 & 6.25 $\pm$ 13.98 & 87.50 $\pm$ 8.84  & 91.88 $\pm$ 6.09 & \textbf{97.50} $\pm$ 2.61\\
    &4m & 20.62 $\pm$ 33.04 & 0.62 $\pm$ 1.4 & 62.50 $\pm$ 21.99 & 73.75 $\pm$ 15.72 & \textbf{75.62 }$\pm$ 21.58\\
    &3m & 6.88 $\pm$ 13.69  & 2.5 $\pm$ 4.07 & 12.50 $\pm$ 18.22 & \textbf{68.75} $\pm$ 17.4 & 64.38 $\pm$ 15.4\\
    \midrule
    \multicolumn{2}{c}{\textbf{Average}} & 48.50 $\pm$ 17.09 & 2.75 $\pm$ 5.19 & 68.38 $\pm$ 11.47 & \textbf{86.88} $\pm$ 4.49 & \textbf{87.50} $\pm$ 6.54\\
\bottomrule
\multicolumn{6}{l}{\small $\diamond$ denotes the source task.}\\
\multicolumn{6}{l}{{\small $^*$ BC-best represents the better per-task and average performance between BC-r and BC-t.}}
\end{tabular}
\end{small}
}
\end{table*}

The One-to-Multi task sets include \textit{Marine-Single} and \textit{Marine-Single-Inv}. Specifically, \textit{Marine-Single} requires agents to train offline using only the dataset from the \textit{3m} task and to test on tasks with scales up to \textit{12m} to assess the agents' generalization ability to larger-scale tasks. (\textit{3m} refers to each side having $3$ Marine units, and similarly for the others). Conversely, \textit{Marine-Single-Inv} requires agents to train offline only from the \textit{10m} task and to test on tasks with a minimum scale of \textit{3m} to evaluate whether the agents can truly learn more general decision-making skills. It is unlikely to expect a well-generalized policy from a small set of tasks with poor datasets. Hence, we use the expert dataset to simulate mastering simpler tasks before extending to more complex ones. {We report the per-task and average results in Table \ref{tab:om-win-rates}, where BC-best denotes the better result between BC-t and BC-r. Full results are presented in Appendix \ref{app:results-otm}.}

{Since} the \textit{Marine-Single} task set only {provides} \textit{3m-expert} data, all algorithms perform better on tasks with smaller agent scales. {However, as the task scale increases, the performance of BC-based methods sharply degrades, whereas CQL-based methods exhibit much less degradation.}
{SD-CQL even maintains} an average win rate of {$27\%$} {on \textit{12m}, which is four times the scale of the source task}. Moreover, SD-CQL exhibits the best performance across all test tasks, with average performance far surpassing other baseline algorithms. {This indicates that SD-CQL acquires general decision-making skills via skill discovery, achieving higher task efficiency and better scalability to unseen tasks.}

For the \textit{Marine-Single-Inv} task set, SD-CQL also demonstrates superior generalization performance on inverse generalization tasks. In contrast, {CQL completely collapses due to error accumulation in tasks with large numbers of agents.} The BC {methods} fail on smaller-scale tasks despite {good performance} on source tasks, as they do not learn generalizable skills. ODIS's skill-learning mechanism alleviates the generalization difficulties inherent in pure BC. However, since its final actions still rely on BC-generated outputs, its performance remains considerably below that of SD-CQL. {The average performance of HiSSD trails only SD-CQL. However, the total number of trainable parameters in HiSSD is more than four times that of ODIS and SD-CQL, further highlighting the simplicity and efficiency of SD-CQL.}

\subsection{Skill-Discovery Visualization}\label{exp_visual}
To provide a more intuitive demonstration of the multi-task generalization capability of SD-CQL, we visualize part of the decision scenarios of the source task \textit{3m} and the target task \textit{12m} from the \textit{Marine-Single} task set in Figure \ref{fig:visual}. Specifically, we record the battle replays (\ref{fig:3m_visual} and \ref{fig:12m_visual}) and project the corresponding skill vectors $z$ at each decision point onto a two-dimensional plane using t-SNE \cite{tsne} (\ref{fig:z_visual}, where each marker represents the skill vector $z$ chosen by an agent at the moment recorded in \ref{fig:3m_visual} and \ref{fig:12m_visual}). More details are available in Appendix \ref{app:visual}. 

It can be observed that SD-CQL successfully learns two skills, \textit{\textbf{retreat}} and \textit{\textbf{attack}}, in the \textit{3m} task and applies them effectively in the unseen \textit{12m} task. When the healthy level is low, the agents choose to \textit{\textbf{retreat}}, while when health is sufficient, they opt to \textit{\textbf{attack}}, which is consistent with the visualization in \ref{fig:z_visual}. This demonstrates that SD-CQL indeed discovers effective skills. Notably, in \textit{12m}, agents adopting the \textit{\textbf{attack}} skill exhibit two specific actions: \textit{firing} and \textit{advancing}, further supporting that SD-CQL extracts cross-task generalizable decision structures instead of mimicking specific actions. {We provide evidence in Appendix \ref{app:visual} that the \textit{advancing} action can be classified as an \textbf{\textit{attack}} skill.}

\section{Conclusion and Future Directions}\label{sec:con}
In this paper, we {address the challenging and underexplored problem of multi-task offline MARL by proposing a novel} task-efficient algorithm, Skill-Discovery Conservative Q-Learning (SD-CQL). It discovers skills offline via {next-observation} reconstruction and trains scalable policies via conservative Q-learning {with local value calibration}. We conduct substantial experiments on StarCraft II to present {the} superior generalization and task-efficiency {of SD-CQL}: It achieves the best performance on {$\textbf{13}$} out of $14$ task sets, with up to {$\textbf{68.9\%}$} improvement on individual task sets. {The results demonstrate} {value-based method} {is not only viable but superior in this setting.}

{Our work suggests several directions for future research, including deriving theoretical insights into the role of skill-discovery in generalization and task efficiency; extending SD-CQL to continuous action spaces; {integrating with alternative skill discovery methods; and enabling offline-to-online fine-tuning on target tasks}. We hope this study encourages further exploration of multi-task offline MARL to advance multi-agent decision-making.} %
\bibliography{plain}

\begin{thebibliography}{10}

\bibitem{IntroCTDE}
Christopher Amato.
\newblock {An} {Introduction} to {Centralized} {Training} for {Decentralized} {Execution} in {Cooperative} {Multi}-{Agent} {Reinforcement} {Learning}, 2024.

\bibitem{chen2024variational}
Jiayu Chen, Tian Lan, and Vaneet Aggarwal.
\newblock Variational offline multi-agent skill discovery.
\newblock {\em arXiv preprint arXiv:2405.16386}, 2024.

\bibitem{de2020independent}
Christian~Schroeder de~Witt, Tarun Gupta, Denys Makoviichuk, Viktor Makoviychuk, Philip~HS Torr, Mingfei Sun, and Shimon Whiteson.
\newblock {Is} independent learning all you need in the starcraft multi-agent challenge?
\newblock {\em arXiv preprint arXiv:2011.09533}, 2020.

\bibitem{fu2020d4rl}
Justin Fu, Aviral Kumar, Ofir Nachum, George Tucker, and Sergey Levine.
\newblock {D}4rl: {D}atasets for deep data-driven reinforcement learning.
\newblock {\em arXiv preprint arXiv:2004.07219}, 2020.

\bibitem{td3bc}
Scott Fujimoto and Shixiang~(Shane) Gu.
\newblock A {Minimalist} {Approach} to {Offline} {Reinforcement} {Learning}.
\newblock In M.~Ranzato, A.~Beygelzimer, Y.~Dauphin, P.S. Liang, and J.~Wortman Vaughan, editors, {\em Advances in Neural Information Processing Systems}, volume~34, pages 20132--20145. Curran Associates, Inc., 2021.

\bibitem{gronauer2022multi}
Sven Gronauer and Klaus Diepold.
\newblock {M}ulti-agent deep reinforcement learning: a survey.
\newblock {\em Artificial Intelligence Review}, 55(2):895--943, 2022.

\bibitem{pymarl2}
Jian Hu, Siying Wang, Siyang Jiang, and Weixun Wang.
\newblock {Rethinking} the {Implementation} {Tricks} and {Monotonicity} {Constraint} in {Cooperative} {Multi}-agent {Reinforcement} {Learning}.
\newblock In {\em ICLR Blogposts 2023}, 2023.
\newblock https://iclr-blogposts.github.io/2023/blog/2023/riit/.

\bibitem{hu2021updet}
Siyi Hu, Fengda Zhu, Xiaojun Chang, and Xiaodan Liang.
\newblock {UPD}e{T}: {U}niversal {M}ulti-agent {RL} via {Policy} {Decoupling} with {Transformers}.
\newblock In {\em International Conference on Learning Representations}, 2021.

\bibitem{jiang2021offline}
Jiechuan Jiang and Zongqing Lu.
\newblock {Offline} decentralized multi-agent reinforcement learning.
\newblock {\em arXiv preprint arXiv:2108.01832}, 2021.

\bibitem{kingma2014adam}
Diederik~P Kingma.
\newblock {A}dam: {A} method for stochastic optimization.
\newblock {\em arXiv preprint arXiv:1412.6980}, 2014.

\bibitem{kostrikov2021offline}
Ilya Kostrikov, Rob Fergus, Jonathan Tompson, and Ofir Nachum.
\newblock {Offline} reinforcement learning with fisher divergence critic regularization.
\newblock In {\em International Conference on Machine Learning}, pages 5774--5783. PMLR, 2021.

\bibitem{kostrikov2021offlines}
Ilya Kostrikov, Ashvin Nair, and Sergey Levine.
\newblock Offline {Reinforcement} {Learning} with {Implicit} {Q}-{Learning}.
\newblock In {\em International Conference on Learning Representations}, 2022.

\bibitem{cql}
Aviral Kumar, Aurick Zhou, George Tucker, and Sergey Levine.
\newblock {Conservative} {Q}-{Learning} for {Offline} {Reinforcement} {Learning}.
\newblock In H.~Larochelle, M.~Ranzato, R.~Hadsell, M.F. Balcan, and H.~Lin, editors, {\em Advances in Neural Information Processing Systems}, volume~33, pages 1179--1191. Curran Associates, Inc., 2020.

\bibitem{lee2007multiagent}
Jae~Won Lee, Jonghun Park, Jangmin O, Jongwoo Lee, and Euyseok Hong.
\newblock A {Multiagent} {Approach} to {$Q$-Learning} for {Daily} {Stock} {Trading}.
\newblock {\em IEEE Transactions on Systems, Man, and Cybernetics - Part A: Systems and Humans}, 37(6):864--877, 2007.

\bibitem{LI2025106852}
Tong Li, Chenjia Bai, Kang Xu, Chen Chu, Peican Zhu, and Zhen Wang.
\newblock {Skill} matters: {Dynamic} skill learning for multi-agent cooperative reinforcement learning.
\newblock {\em Neural Networks}, 181:106852, 2025.

\bibitem{li2023beyond}
Zhuoran Li, Ling Pan, and Longbo Huang.
\newblock {B}eyond conservatism: {D}iffusion policies in offline multi-agent reinforcement learning.
\newblock {\em arXiv preprint arXiv:2307.01472}, 2023.

\bibitem{lin2024hgap}
Bor-Jiun Lin and Chun-Yi Lee.
\newblock {HGAP}: Boosting permutation invariant and permutation equivariant in multi-agent reinforcement learning via graph attention network.
\newblock In {\em Forty-first International Conference on Machine Learning}, 2024.

\bibitem{pmlr-v100-liu20a}
Iou-Jen Liu, Raymond~A. Yeh, and Alexander~G. Schwing.
\newblock Pic: Permutation invariant critic for multi-agent deep reinforcement learning.
\newblock In Leslie~Pack Kaelbling, Danica Kragic, and Komei Sugiura, editors, {\em Proceedings of the Conference on Robot Learning}, volume 100 of {\em Proceedings of Machine Learning Research}, pages 590--602. PMLR, 30 Oct--01 Nov 2020.

\bibitem{liu2024maximum}
Jiarong Liu, Yifan Zhong, Siyi Hu, Haobo Fu, QIANG FU, Xiaojun Chang, and Yaodong Yang.
\newblock {Maximum} {Entropy} {Heterogeneous}-{Agent} {Reinforcement} {Learning}.
\newblock In {\em The Twelfth International Conference on Learning Representations}, 2024.

\bibitem{hissd}
Sicong Liu, Yang Shu, Chenjuan Guo, and Bin Yang.
\newblock {Learning} {Generalizable} {Skills} from {Offline} {Multi}-{Task} {Data} for {Multi}-{Agent} {Cooperation}.
\newblock In {\em The Thirteenth International Conference on Learning Representations}, 2025.

\bibitem{hsl2022}
Yuntao Liu, Yuan Li, Xinhai Xu, Yong Dou, and Donghong Liu.
\newblock {Heterogeneous} {Skill} {Learning} for {Multi}-agent {Tasks}.
\newblock In S.~Koyejo, S.~Mohamed, A.~Agarwal, D.~Belgrave, K.~Cho, and A.~Oh, editors, {\em Advances in Neural Information Processing Systems}, volume~35, pages 37011--37023. Curran Associates, Inc., 2022.

\bibitem{liu2024offlinemultiagentreinforcementlearning}
Zongkai Liu, Qian Lin, Chao Yu, Xiawei Wu, Yile Liang, Donghui Li, and Xuetao Ding.
\newblock {Offline} {Multi}-{Agent} {Reinforcement} {Learning} via {In}-{Sample} {Sequential} {Policy} {Optimization}, 2024.

\bibitem{lowe2017multi}
Ryan Lowe, YI~WU, Aviv Tamar, Jean Harb, OpenAI Pieter~Abbeel, and Igor Mordatch.
\newblock Multi-agent actor-critic for mixed cooperative-competitive environments.
\newblock In I.~Guyon, U.~Von Luxburg, S.~Bengio, H.~Wallach, R.~Fergus, S.~Vishwanathan, and R.~Garnett, editors, {\em Advances in Neural Information Processing Systems}, volume~30. Curran Associates, Inc., 2017.

\bibitem{mathieu2021starcraft}
Michael Mathieu, Sherjil Ozair, Srivatsan Srinivasan, Caglar Gulcehre, Shangtong Zhang, Ray Jiang, Tom~Le Paine, Konrad Zolna, Richard Powell, Julian Schrittwieser, David Choi, Petko Georgiev, Daniel~Kenji Toyama, Aja Huang, Roman Ring, Igor Babuschkin, Timo Ewalds, Mahyar Bordbar, Sarah Henderson, Sergio~G{\'o}mez Colmenarejo, Aaron van~den Oord, Wojciech~M. Czarnecki, Nando de~Freitas, and Oriol Vinyals.
\newblock {StarCraft} {II} {Unplugged}: {Large} {Scale} {Offline} {Reinforcement} {Learning}.
\newblock In {\em Deep RL Workshop NeurIPS 2021}, 2021.

\bibitem{matignon2012coordinated}
Laetitia Matignon, Laurent Jeanpierre, and Abdel-Illah Mouaddib.
\newblock {Coordinated} {Multi}-{Robot} {Exploration} {Under} {Communication} {Constraints} {Using} {Decentralized} {Markov} {Decision} {Processes}.
\newblock {\em Proceedings of the AAAI Conference on Artificial Intelligence}, 26(1):2017--2023, Sep. 2021.

\bibitem{oliehoek2016concise}
Frans~A Oliehoek, Christopher Amato, et~al.
\newblock {\em {A} concise introduction to decentralized {POMDP}s}, volume~1.
\newblock Springer, 2016.

\bibitem{oliehoek2008optimal}
Frans~A Oliehoek, Matthijs~TJ Spaan, and Nikos Vlassis.
\newblock {O}ptimal and approximate {Q}-value functions for decentralized {POMDPs}.
\newblock {\em Journal of Artificial Intelligence Research}, 32:289--353, 2008.

\bibitem{pmlr-v70-omidshafiei17a}
Shayegan Omidshafiei, Jason Pazis, Christopher Amato, Jonathan~P. How, and John Vian.
\newblock Deep decentralized multi-task multi-agent reinforcement learning under partial observability.
\newblock In Doina Precup and Yee~Whye Teh, editors, {\em Proceedings of the 34th International Conference on Machine Learning}, volume~70 of {\em Proceedings of Machine Learning Research}, pages 2681--2690. PMLR, 06--11 Aug 2017.

\bibitem{omar}
Ling Pan, Longbo Huang, Tengyu Ma, and Huazhe Xu.
\newblock {Plan} {Better} {Amid} {Conservatism}: {Offline} {Multi}-{Agent} {Reinforcement} {Learning} with {Actor} {Rectification}.
\newblock In Kamalika Chaudhuri, Stefanie Jegelka, Le~Song, Csaba Szepesvari, Gang Niu, and Sivan Sabato, editors, {\em Proceedings of the 39th International Conference on Machine Learning}, volume 162 of {\em Proceedings of Machine Learning Research}, pages 17221--17237. PMLR, 17--23 Jul 2022.

\bibitem{res}
Ling Pan, Tabish Rashid, Bei Peng, Longbo Huang, and Shimon Whiteson.
\newblock Regularized {Softmax} {Deep} {Multi}-{Agent} {Q}-{Learning}.
\newblock In M.~Ranzato, A.~Beygelzimer, Y.~Dauphin, P.S. Liang, and J.~Wortman Vaughan, editors, {\em Advances in Neural Information Processing Systems}, volume~34, pages 1365--1377. Curran Associates, Inc., 2021.

\bibitem{qmix}
Tabish Rashid, Mikayel Samvelyan, Christian~Schroeder de~Witt, Gregory Farquhar, Jakob Foerster, and Shimon Whiteson.
\newblock {Monotonic} {Value} {Function} {Factorisation} for {Deep} {Multi}-{Agent} {Reinforcement} {Learning}.
\newblock {\em Journal of Machine Learning Research}, 21(178):1--51, 2020.

\bibitem{rezaeifar2022offline}
Shideh Rezaeifar, Robert Dadashi, Nino Vieillard, Léonard Hussenot, Olivier Bachem, Olivier Pietquin, and Matthieu Geist.
\newblock {Offline} {Reinforcement} {Learning} as {Anti}-exploration.
\newblock {\em Proceedings of the AAAI Conference on Artificial Intelligence}, 36(7):8106--8114, Jun. 2022.

\bibitem{samvelyan19smac}
Mikayel Samvelyan, Tabish Rashid, Christian Schroeder~de Witt, Gregory Farquhar, Nantas Nardelli, Tim G.~J. Rudner, Chia-Man Hung, Philip H.~S. Torr, Jakob Foerster, and Shimon Whiteson.
\newblock {The} {StarCraft} {Multi}-{Agent} {Challenge}.
\newblock In {\em Proceedings of the 18th International Conference on Autonomous Agents and MultiAgent Systems}, AAMAS '19, page 2186–2188, Richland, SC, 2019. International Foundation for Autonomous Agents and Multiagent Systems.

\bibitem{shalev2016safe}
Shai Shalev-Shwartz, Shaked Shammah, and Amnon Shashua.
\newblock {S}afe, multi-agent, reinforcement learning for autonomous driving.
\newblock {\em arXiv preprint arXiv:1610.03295}, 2016.

\bibitem{shao2023counterfactual}
Jianzhun Shao, Yun Qu, Chen Chen, Hongchang Zhang, and Xiangyang Ji.
\newblock {Counterfactual} {Conservative} {Q} {Learning} for {Offline} {Multi}-agent {Reinforcement} {Learning}.
\newblock In {\em Thirty-seventh Conference on Neural Information Processing Systems}, 2023.

\bibitem{shoham2008multiagent}
Yoav Shoham and Kevin Leyton-Brown.
\newblock {\em {M}ultiagent systems: {A}lgorithmic, game-theoretic, and logical foundations}.
\newblock Cambridge University Press, 2008.

\bibitem{stone2010ad}
Peter Stone, Gal Kaminka, Sarit Kraus, and Jeffrey Rosenschein.
\newblock Ad hoc autonomous agent teams: Collaboration without pre-coordination.
\newblock {\em Proceedings of the AAAI Conference on Artificial Intelligence}, 24(1):1504--1509, Jul. 2010.

\bibitem{sunehag2018value}
Peter Sunehag, Guy Lever, Audrunas Gruslys, Wojciech~Marian Czarnecki, Vinicius Zambaldi, Max Jaderberg, Marc Lanctot, Nicolas Sonnerat, Joel~Z. Leibo, Karl Tuyls, and Thore Graepel.
\newblock {Value}-{Decomposition} {Networks} {For} {Cooperative} {Multi}-{Agent} {Learning} {Based} {On} {Team} {Reward}.
\newblock In {\em Proceedings of the 17th International Conference on Autonomous Agents and MultiAgent Systems}, AAMAS '18, page 2085–2087, Richland, SC, 2018. International Foundation for Autonomous Agents and Multiagent Systems.

\bibitem{tampuu2017multiagent}
Ardi Tampuu, Tambet Matiisen, Dorian Kodelja, Ilya Kuzovkin, Kristjan Korjus, Juhan Aru, Jaan Aru, and Raul Vicente.
\newblock Multiagent cooperation and competition with deep reinforcement learning.
\newblock {\em PLOS ONE}, 12(4):1--15, 04 2017.

\bibitem{tian2024decompose}
Zikang Tian, Ruizhi Chen, Xing Hu, Ling Li, Rui Zhang, Fan Wu, Shaohui Peng, Jiaming Guo, Zidong Du, Qi~Guo, and Yunji Chen.
\newblock {Decompose} a {Task} into {Generalizable} {Subtasks} in {Multi}-{Agent} {Reinforcement} {Learning}.
\newblock In A.~Oh, T.~Naumann, A.~Globerson, K.~Saenko, M.~Hardt, and S.~Levine, editors, {\em Advances in Neural Information Processing Systems}, volume~36, pages 78514--78532. Curran Associates, Inc., 2023.

\bibitem{tsne}
Laurens Van~der Maaten and Geoffrey Hinton.
\newblock {V}isualizing data using t-{SNE}.
\newblock {\em Journal of machine learning research}, 9(11), 2008.

\bibitem{vaswani2017attention}
A~Vaswani.
\newblock {Attention} is all you need.
\newblock {\em Advances in Neural Information Processing Systems}, 2017.

\bibitem{wang2020rode}
Tonghan Wang, Tarun Gupta, Anuj Mahajan, Bei Peng, Shimon Whiteson, and Chongjie Zhang.
\newblock {RODE}: {Learning} {Roles} to {Decompose} {Multi}-{Agent} {Tasks}.
\newblock In {\em International Conference on Learning Representations}, 2021.

\bibitem{wang2020few}
Weixun Wang, Tianpei Yang, Yong Liu, Jianye Hao, Xiaotian Hao, Yujing Hu, Yingfeng Chen, Changjie Fan, and Yang Gao.
\newblock From few to more: Large-scale dynamic multiagent curriculum learning.
\newblock {\em Proceedings of the AAAI Conference on Artificial Intelligence}, 34(05):7293--7300, Apr. 2020.

\bibitem{wu2024portal}
Jizhou Wu, Jianye Hao, Tianpei Yang, Xiaotian Hao, Yan Zheng, Weixun Wang, and Matthew~E. Taylor.
\newblock {PORTAL}: {Automatic} {Curricula} {Generation} for {Multiagent} {Reinforcement} {Learning}.
\newblock {\em Proceedings of the AAAI Conference on Artificial Intelligence}, 38(14):15934--15942, Mar. 2024.

\bibitem{HMASD}
Mingyu Yang, Yaodong Yang, Zhenbo Lu, Wengang Zhou, and Houqiang Li.
\newblock {Hierarchical} {Multi}-{Agent} {Skill} {Discovery}.
\newblock In A.~Oh, T.~Naumann, A.~Globerson, K.~Saenko, M.~Hardt, and S.~Levine, editors, {\em Advances in Neural Information Processing Systems}, volume~36, pages 61759--61776. Curran Associates, Inc., 2023.

\bibitem{yang2021believe}
Yiqin Yang, Xiaoteng Ma, Chenghao Li, Zewu Zheng, Qiyuan Zhang, Gao Huang, Jun Yang, and Qianchuan Zhao.
\newblock {Believe} {What} {You} {See}: {Implicit} {Constraint} {Approach} for {Offline} {Multi}-{Agent} {Reinforcement} {Learning}.
\newblock In M.~Ranzato, A.~Beygelzimer, Y.~Dauphin, P.S. Liang, and J.~Wortman Vaughan, editors, {\em Advances in Neural Information Processing Systems}, volume~34, pages 10299--10312. Curran Associates, Inc., 2021.

\bibitem{yu2021surprising}
Chao Yu, Akash Velu, Eugene Vinitsky, Jiaxuan Gao, Yu~Wang, Alexandre Bayen, and YI~WU.
\newblock The {Surprising} {Effectiveness} of {PPO} in {Cooperative} {Multi}-{Agent} {Games}.
\newblock In S.~Koyejo, S.~Mohamed, A.~Agarwal, D.~Belgrave, K.~Cho, and A.~Oh, editors, {\em Advances in Neural Information Processing Systems}, volume~35, pages 24611--24624. Curran Associates, Inc., 2022.

\bibitem{zhang2023odis}
Fuxiang Zhang, Chengxing Jia, Yi-Chen Li, Lei Yuan, Yang Yu, and Zongzhang Zhang.
\newblock {Discovering} {Generalizable} {Multi}-agent {Coordination} {Skills} from {Multi}-task {Offline} {Data}.
\newblock In {\em The Eleventh International Conference on Learning Representations}, 2023.

\bibitem{zhang2023fast}
Ziqian Zhang, Lei Yuan, Lihe Li, Ke~Xue, Chengxing Jia, Cong Guan, Chao Qian, and Yang Yu.
\newblock {Fast} {Teammate} {Adaptation} in the {Presence} of {Sudden} {Policy} {Change}.
\newblock In Robin~J. Evans and Ilya Shpitser, editors, {\em Proceedings of the Thirty-Ninth Conference on Uncertainty in Artificial Intelligence}, volume 216 of {\em Proceedings of Machine Learning Research}, pages 2465--2476. PMLR, 31 Jul--04 Aug 2023.

\end{thebibliography}
\bibliographystyle{plain}

\newpage
\appendix
\renewcommand{\appendixpagename}{\LARGE Technical Appendices}
\appendixpage
\startcontents[section]
\printcontents[section]{l}{1}{\setcounter{tocdepth}{2}}

\newpage

\section{Related Work} \label{app:rw}
\paragraph{MARL}
Multi-Agent Reinforcement Learning (MARL) has seen substantial progress in recent years, with numerous approaches developed under different paradigms. 
The centralized training with decentralized execution (CTDE) paradigm \cite{oliehoek2008optimal,matignon2012coordinated,IntroCTDE} has been particularly influential, with methods such as HASAC \cite{liu2024maximum}, MAPPO \cite{yu2021surprising}, QMIX \cite{qmix}, VDN \cite{sunehag2018value}, and MADDPG \cite{lowe2017multi}. These approaches use centralized training for better coordination and decentralized execution for real-time decision-making. On the other hand, fully decentralized training and execution schemes have also been explored \cite{tampuu2017multiagent,de2020independent}. However, the performance of such methods is often constrained by the absence of information sharing. In this paper, we mainly focus on the CTDE paradigm with QMIX as the backbone.

\paragraph{Offline MARL}
Due to the absence of online interaction with the environment, offline training faces a fundamental challenge—distributional shift. To address this issue, several techniques for single-agent RL have been proposed, many of which leverage conservatism to regularize either the policy \cite{td3bc, kostrikov2021offline} or the Q-value function \cite{cql,kostrikov2021offlines,rezaeifar2022offline}. These methods mitigate the risks of overestimating the value of unseen state-action pairs. However, specific challenges caused by multiple agents, such as the exponential explosion of complexity, hinder these techniques from directly extending to multi-agent scenarios. Therefore, several tailored approaches \cite{jiang2021offline, yang2021believe, omar, li2023beyond,  shao2023counterfactual, liu2024offlinemultiagentreinforcementlearning} have been proposed to address offline MARL. However, these methods often focus excessively on source tasks, inevitably compromising their multi-task generalization ability.

\paragraph{Multi-task MARL} 
Current research on multi-task MARL can be broadly categorized into two types. The first, often referred to as Ad-Hoc Teamwork \cite{stone2010ad, zhang2023fast}, focuses on exploring how to effectively collaborate with unknown and uncontrollable teammates within a given task. The second type, which is the primary focus of this paper, involves scenarios where the algorithm controls the entire team, but the agents are trained on some tasks and tested on unseen tasks with similar structures \cite{pmlr-v70-omidshafiei17a,wang2020few, wu2024portal}. This requires agents to learn and utilize generalizable decision-making structures across tasks from a limited set of source tasks, positioning skill discovery as a promising solution. However, most existing approaches discover skills typically through sample reconstruction \cite{wang2020rode,hsl2022,HMASD}, clustering \cite{LI2025106852}, or sub-task decomposition \cite{tian2024decompose}, primarily relying on online interactions, with limited attention to offline settings. {VO-MASD \cite{chen2024variational} explores offline skill discovery with a codebook. However, its emphasis lies in training high-level policies online using the discovered skills, rather than in a fully offline setting.}  Recently, ODIS \cite{zhang2023odis} proposed a method for multi-task offline MARL, but it suffers from the limitations of behavior cloning when the scale of source tasks is small. {Inspired by ODIS, HiSSD \cite{hissd} leverages value functions to separately extract common and task-specific skills, improving generalization performance. However, its task efficiency remains constrained by the BC action decoder, which hinders the discovery of effective skills from a limited number of source tasks.}

\section{Algorithm Pseudocode}\label{app:algo}

\begin{algorithm}[!htpb]
   \caption{Skill-Discovery Conservative Q-Learning}
   \label{alg:sdcql}
\begin{algorithmic}
   \STATE {\bfseries Input:} Datasets of source tasks $\{\mathcal{D}_m\}_{m\in\mathbb{T}_s}$.
   \STATE Initialize the parameters $\phi$ for the encoder and decoder, $\theta$ for $Q_{tot}$ and $\bar\theta$ for target $Q_{tot}$.
   \FOR{$i=1${\bfseries to} $T_{\text{max}}$}
   \FOR{$j=1$ {\bfseries to} $|\mathbb{T}_s|$}
   \STATE Sample trajectories $\hat\tau=\{(s_t,o_t,a_t,r_t,o'_t)\}$ from $\mathcal{D}_j$.
   \STATE Calculate the total loss $\mathcal{L}_{j}$ by $\hat\tau$ for task $j$, according to \eqref{total_loss}.
   \ENDFOR
   \STATE Calculate the multi-task loss $\mathcal{L}=\sum_{j=1}^{|\mathbb{T}_s|}\mathcal{L}_{j}$.
   \STATE Update $\phi$ with $\nabla_\phi \mathcal{L}$ and update $\theta$ with $\nabla_\theta \mathcal{L}$
   \IF{$i \equiv 0 \pmod {T_{\text{update}}}$}
   \STATE $\bar\theta \leftarrow \theta$
   \ENDIF
   \ENDFOR
\end{algorithmic}
\end{algorithm}

\section{Experiment Details}\label{app:exp_detail}
\subsection{Datasets}\label{app:datasets}
Following the definition by \cite{fu2020d4rl}, we conduct experiments primarily on datasets of four qualities. The collection procedure for each quality is as follows: 
\begin{itemize}
    \setlength{\itemsep}{0pt}
    \item Expert: Trajectories collected from policies trained to an expert level using QMIX.
    \item Medium: Trajectories collected from policies trained to a medium level using QMIX.
    \item Medium-Replay: The replay buffer generated while training QMIX policies to a medium level.
    \item Medium-Expert: A 50-50 mixture of Medium and Expert trajectories.
\end{itemize}
To ensure fair comparisons, we use the datasets provided by ODIS \cite{zhang2023odis}, where only up to $2,000$ trajectories are sampled for each dataset. The key information of the datasets used in our experiments is summarized in Table \ref{tab: datasets}.

\begin{table}[!ht]
\caption{The primary information of the offline datasets used in our experiments.}
\label{tab: datasets}
\begin{center}
\begin{small}
\begin{tabular}{lcc}
    \toprule
    \textbf{Task} & \textbf{\#Trajectories} & \textbf{Average Return}\\
    \midrule
    3m-Expert & 2,000 & 19.87\\
    3m-Medium & 2,000 & 14.00\\
    3m-Medium-Replay & 2,000 & 10.71\\
    3m-Medium-Expert & 2,000 & 16.94\\
    5m-Expert & 2,000 & 19.93\\
    5m-Medium & 2,000 & 17.35\\
    5m-Medium-Replay & 1,266 & 3.21\\
    5m-Medium-Expert & 2,000 & 18.66\\
    10m-Expert & 2,000 & 19.89\\
    10m-Medium & 2,000 & 16.63\\
    10m-Medium-Replay & 331 & 2.30\\
    10m-Medium-Expert & 2,000 & 18.27\\
    \midrule
    5m\_vs\_6m-Expert & 2,000 & 17.34\\
    5m\_vs\_6m-Medium & 2,000 & 12.63\\
    5m\_vs\_6m-Medium-Replay & 2,000 & 9.41\\
    5m\_vs\_6m-Medium-Expert & 2,000 & 14.98\\
    9m\_vs\_10m-Expert & 2,000 & 19.59\\
    9m\_vs\_10m-Medium & 2,000 & 15.52\\
    9m\_vs\_10m-Medium-Replay & 2,000 & 11.76\\
    9m\_vs\_10m-Medium-Expert & 2,000 & 17.49\\
    \midrule
    2s3z-Expert & 2,000 & 19.78\\
    2s3z-Medium & 2,000 & 16.61\\
    2s3z-Medium-Replay & 2,000 & 14.25\\
    2s3z-Medium-Expert & 2,000 & 18.26\\
    2s4z-Expert & 2,000 & 19.73\\
    2s4z-Medium & 2,000 & 16.85\\
    2s4z-Medium-Replay & 2,000 & 14.38\\
    2s4z-Medium-Expert & 2,000 & 18.26\\
    3s5z-Expert & 2,000 & 19.78\\
    3s5z-Medium & 2,000 & 16.31\\
    3s5z-Medium-Replay & 2,000 & 15.29\\
    3s5z-Medium-Expert & 2,000 & 18.05\\
    \bottomrule
\end{tabular}
\end{small}
\end{center}
\end{table}
\newpage
\subsection{Tasks Configuration}\label{app:tasks}
In this section, we present the key information of all tasks involved in our experiments in Table \ref{tab: tasks} and the task composition of each task set in Table \ref{tab: tasksets}.
\begin{table}[!ht]
\caption{The key information about the tasks used in our experiments.}
\label{tab: tasks}
\begin{center}
\begin{small}
\begin{tabular}{lll}
    \toprule
    \textbf{Task Name} & \textbf{Allied Units}& \textbf{Enemy Units}\\
    \midrule
    3m & 3 Marines & 3 Marines \\
    4m & 4 Marines & 4 Marines \\
    5m & 5 Marines & 5 Marines \\
    6m & 6 Marines & 6 Marines \\
    7m & 7 Marines & 7 Marines \\
    8m & 8 Marines & 8 Marines \\
    9m & 9 Marines & 9 Marines \\
    10m & 10 Marines & 10 Marines \\
    11m & 11 Marines & 11 Marines \\
    12m & 12 Marines & 12 Marines \\
    \midrule
    5m\_vs\_6m & 5 Marines & 6 Marines \\
    6m\_vs\_7m & 6 Marines & 7 Marines \\
    7m\_vs\_8m & 7 Marines & 8 Marines \\
    8m\_vs\_9m & 8 Marines & 9 Marines \\
    9m\_vs\_10m & 9 Marines & 10 Marines \\
    10m\_vs\_11m & 10 Marines & 11 Marines \\
    10m\_vs\_12m & 10 Marines & 12 Marines \\
    13m\_vs\_15m & 13 Marines & 15 Marines \\
    \midrule
    1s3z & 1 Stalkers, 3 Zealots & 1 Stalkers, 3 Zealots \\
    1s4z & 1 Stalkers, 4 Zealots & 1 Stalkers, 4 Zealots \\
    1s5z & 1 Stalkers, 5 Zealots & 1 Stalkers, 5 Zealots \\
    2s3z & 2 Stalkers, 3 Zealots & 2 Stalkers, 3 Zealots \\
    2s4z & 2 Stalkers, 4 Zealots & 2 Stalkers, 4 Zealots \\
    2s5z & 2 Stalkers, 5 Zealots & 2 Stalkers, 5 Zealots \\
    3s3z & 3 Stalkers, 3 Zealots & 3 Stalkers, 3 Zealots \\
    3s4z & 3 Stalkers, 4 Zealots & 3 Stalkers, 4 Zealots \\
    3s5z & 3 Stalkers, 5 Zealots & 3 Stalkers, 5 Zealots \\
    4s3z & 4 Stalkers, 3 Zealots & 4 Stalkers, 3 Zealots \\
    4s4z & 4 Stalkers, 4 Zealots & 4 Stalkers, 4 Zealots \\
    4s5z & 4 Stalkers, 5 Zealots & 4 Stalkers, 5 Zealots \\
    4s6z & 4 Stalkers, 6 Zealots & 4 Stalkers, 6 Zealots \\
    \bottomrule
\end{tabular}
\end{small}
\end{center}
\end{table}

\begin{table}[!ht]
\caption{The task composition of each {scenario} used in our experiments.}
\label{tab: tasksets}
\begin{center}
\begin{small}
\begin{tabular}{lll}
    \toprule
    \textbf{{Scenarios}} & \textbf{Source Tasks}& \textbf{Unseen Tasks}\\
    \midrule
    Marine-Easy & 3m, 5m, 10m & 4m, 6m, 7m, 8m, 9m, 11m, 12m \\
    \midrule
    \multirow{2}{*}{Marine-Hard} & \multirow{2}{*}{3m, 5m\_vs\_6m, 9m\_vs\_10m} & 4m, 5m, 10m, 12m, \\
    & & 6m\_vs\_7m, 7m\_vs\_8m, 8m\_vs\_9m, \\
    & & 10m\_vs\_11m, 10m\_vs\_12m, \\
    & & 13m\_vs\_15m \\
    \midrule
    Stalker-Zealot & 2s3z, 2s4z, 3s5z & 1s3z, 1s4z, 1s5z, 2s5z, 3s3z, 3s4z,\\
    & &  4s3z, 4s4z, 4s5z, 4s6z \\
    \midrule
    Marine-Single & 3m & 5m, 8m, 10m, 12m \\
    \midrule
    Marine-Single-Inv & 10m & 3m, 4m, 5m, 8m \\
    \bottomrule
\end{tabular}
\end{small}
\end{center}
\end{table}

\newpage

\subsection{Implementation Details}\label{app:imp_detail}
\subsubsection{Decomposer}
{To handle variable-sized inputs across different tasks, we follow prior works \cite{hu2021updet,tian2024decompose,zhang2023odis, hissd} and adopt a rule-based decomposer to split the input observations or states into individual entity information units. This is feasible because SMAC provides a well-defined and consistent structure for states and observations across tasks.}

{Specifically, taking observations as an example, each agent’s observation is a concatenation of its own features, the features of all allies, and the features of all enemies. While the feature schema for each entity is fixed and standardized, the number of allies and enemies depends on the task configuration. For entities that are not observable (e.g., outside the agent's field of view), their corresponding feature fields are zero-masked. This allows a deterministic rule-based decomposer to segment the observation into individual entity features based on the known observation layout and the number of agents involved in the current task.}

{Importantly, the decomposer operates solely based on environment-level specifications rather than task-level ones. And, as discussed in the main text, our study focuses on a series of structurally similar tasks that differ in scale, difficulty, and objectives. Therefore, although SMAC explicitly provides the number of interactable entities and the total number of other entities, similar information can be obtained in other environments using techniques such as entity detection or masking mechanisms. Hence, this component does not limit the generality or contribution of SD-CQL.}

\subsubsection{SD-CQL}
\paragraph{{Skill Discovery.}} In the {skill discovery} part, we use single-layer Transformers {for both the encoder and the decoder}, and fully connected layers for embedding and skill extraction. {To make the skill vector extract more information, the decoder reconstructs the embedding associated with the agent itself directly with $z$:}
$$
{\widehat{E}_{i,0} = W_0 \cdot \text{ReLU}(z) + b_0}.
$$ 
{While for embeddings of other entities, it first masks parts of the information by a ReLU function, and then concatenates them with $z$ for reconstruction:}
\begin{equation}
{\widehat{E}_{i,k} = W_k \cdot \left[\text{ReLU}(E_{i,k}),~z\right] + b_k,\quad k=1,2,\dots,K_i}
\end{equation}
{where $[\cdot,~\cdot]$ represents the concatenation operation.}

\paragraph{{Skill-conditioned Policy Optimization.}} In the {skill-conditioned policy optimization} part, we use MLPs as the individual Q network, and similar to ODIS, we employ an attention-based mixing network to handle variable inputs in multi-task training. The entity decomposition method and its dimensions are the same as those used in the observation reconstruction. The specific parameters of the network structure are shown in Table \ref{tab: arch}. For Conservative Q-Learning, the sampling strategy $\mu$ we use is a uniform distribution over the action space, with 100 samples drawn to estimate the CQL regularization term. Additionally, consistent with the QMIX implementation provided by PYMARL2 \cite{pymarl2}, we use TD($\lambda$) to enhance stability when calculating the TD loss, and the $\lambda$ parameter is set to the default value from PYMARL2.

\begin{table}[!ht]
\caption{The main network structure involved in SD-CQL.}
\label{tab: arch}
\begin{center}
\begin{small}
\begin{tabular}{lccc}
    \toprule
    \textbf{Network} & \textbf{Architecture} & \textbf{Activation Function}\\
    \midrule
    Transformer & Head=1, Depth=1, Embedding Dim=64& ——\\
    Fixed Individual Q Network& $[96,128,6]$ & ReLU\\
    Variable Individual Q Network &$[96,128,1]$& ReLU\\
    HyperNet in Mixing Network& $[128,64,32]$& ReLU\\
    Attention in Mixing Network& Attention Dim=$8$ & ——\\
    \bottomrule
\end{tabular}
\end{small}
\end{center}
\end{table}

\subsubsection{Baselines}
For the baselines {except for HiSSD}, we utilize the implementations provided by ODIS. Specifically, BC-t employs a transformer capable of decomposing observations for multi-task training while optimizing the policy through behavior cloning. BC-r is similar to BC-t but incorporates return-to-go information as an additional input alongside observations. {The vanilla CQL is almost the same as SD-CQL, but it removes the skill-discovery module and the local value calibration term. It is only equipped with the observation encoder, and feeds the encoding vectors} ${\{E_{i,k}\}_{k=0}^{K_i}}$ {to separated Q-networks directly. Additionally, in vanilla CQL, the gradient can propagate from the encoder to the Q-networks. }{The implementations of ODIS and HiSSD align with their official version.}

\subsection{Hyperparameters}\label{app:hyperparam}
For SD-CQL, the main hyperparameters shared across all tasks are listed in Table \ref{tab: hyper}. We primarily adjust the {local value calibration} coefficient $\eta$, with the specific settings provided in Table \ref{tab: bchyper}. All other hyperparameters are kept consistent with the default configuration of QMIX in PYMARL2.

\begin{table}[!htbp]
\caption{The hyperparameters shared by all tasks for SD-CQL.}
\label{tab: hyper}
\begin{center}
\begin{small}
\begin{tabular}{lcc}
    \toprule
    \textbf{Hyperparameter} & \textbf{Value}\\
    \midrule
    Entity Embedding Dim & 64 \\
    Skill Dimension of $z$ & 32 \\
    $\lambda$ for $TD(\lambda)$ & 0.6\\
    Conservative Weight $\alpha$ & 1.0\\
    $T_{\text{max}}$ & 35,000\\
    $T_{\text{update}}$ & 80\\
    Batch Size & 32 \\
    Learning Rate & 0.005\\
    Optimizer & Adam \cite{kingma2014adam} \\
    \bottomrule
\end{tabular}
\end{small}
\end{center}
\end{table}

\begin{table}[!htbp]
\caption{{Local value calibration} coefficient $\eta$ for each task set.}
\label{tab: bchyper}
\begin{center}
\begin{small}
\begin{tabular}{lcc}
    \toprule
    \textbf{Task set} & \textbf{Value}\\
    \midrule
    Marine-Easy-Expert & 0.9 \\
    Marine-Easy-Medium & 0.2 \\
    Marine-Easy-Medium-Replay & 0.5 \\
    Marine-Easy-Medium-Expert & 0.5 \\
    \midrule
    Marine-Hard-Expert & 0.5 \\
    Marine-Hard-Medium & 0.8 \\
    Marine-Hard-Medium-Replay & 0.9 \\
    Marine-Hard-Medium-Expert & 0.7 \\
    \midrule
    Stalker-Zealot-Expert & 0.3 \\
    Stalker-Zealot-Medium & 0.1 \\
    Stalker-Zealot-Medium-Replay & 0.5 \\
    Stalker-Zealot-Medium-Expert & 0.3 \\
    \midrule
    Marine-Single & 0.5 \\
    Marine-Single-Inv & 0.3 \\
    \bottomrule
\end{tabular}
\end{small}
\end{center}
\end{table}

Since the \textit{Marine-Single} and \textit{Marine-Single-Inv} task sets designed by us, we adjust {the primary hyperparameters involved in the RL losses of ODIS and HiSSD. For ODIS, we tune the conservative coefficient $\alpha$ (selected from $\{1.0, 2.5, 5.0\}$) and the distribution coefficient $\lambda$ (from $\{1.0, 2.5, 5.0\}$). For HiSSD, due to training on only one source task, the contrastive loss cannot be applied. Consequently, we only tune the planner weight $\alpha$ (selected from $\{5.0, 10.0, 20.0\}$) and the expectile regression loss parameter $\epsilon$ (from $\{0.5, 0.7, 0.9\}$). }

{We select the best configuration of each for reporting. The final hyperparameter settings are provided in Table \ref{tab: ohhyper}, while Table \ref{tab: odisperf} and Table \ref{tab: hissdperf} present the average performance of all configurations we explored.}

\begin{table}[!ht]
\caption{Selected Hyperparameters for ODIS and HiSSD.}
\label{tab: ohhyper}
\begin{center}
\begin{small}
\begin{tabular}{lcc}
    \toprule
    \textbf{Task set} & \textbf{ODIS} & \textbf{HiSSD}\\
    \midrule
    \textit{Marine-Single} & $\alpha=5.0$, $\lambda=1.0$ & $\alpha=10.0$, $\epsilon=0.9$\\
    \textit{Marine-Single-Inv} & $\alpha=1.0$, $\lambda=1.0$ & $\alpha=5.0$, $\epsilon=0.7$\\
    \bottomrule
\end{tabular}
\end{small}
\end{center}
\end{table}

\begin{table}[!ht]
\caption{Average win rates of all configurations explored for ODIS. The reported results are averaged over 5 random seeds. The bold number denotes the best performance, and $\pm$ denotes one standard deviation.}
\label{tab: odisperf}
\begin{center}
\begin{small}
\begin{tabular}{cccc}
    \toprule
    $\bm\alpha$ & $\bm\lambda$ & \textbf{Marine-Single} & \textbf{Marine-Single-Inv} \\
    \midrule
    $1.0$& $1.0$ & 27.12 $\pm$ 7.39  & \textbf{68.38} $\pm$ 11.47  \\
    $1.0$& $2.5$ & 26.88 $\pm$ 4.56  &  62.88 $\pm$ 12.78 \\
    $1.0$& $5.0$ & 23.38 $\pm$ 3.30  & 53.25 $\pm$ 13.59  \\
    $2.5$& $1.0$ & 31.38 $\pm$ 8.55  & 61.50 $\pm$ 13.06  \\
    $2.5$& $2.5$ & 25.25 $\pm$ 5.99  &  64.12 $\pm$ 14.06 \\
    $2.5$& $5.0$ & 25.25 $\pm$ 2.11  &  64.00 $\pm$ 2.64 \\
    $5.0$& $1.0$ & \textbf{33.25} $\pm$ 16.50  & 66.12 $\pm$ 8.73   \\
    $5.0$& $2.5$ & 22.25 $\pm$ 1.29  & 55.62 $\pm$ 12.23  \\
    $5.0$& $5.0$ & 24.62 $\pm$ 4.93  &  52.12 $\pm$ 17.44	 \\
    \bottomrule
\end{tabular}
\end{small}
\end{center}
\end{table}

\begin{table}[!htbp]
\caption{Average win rates of all configurations explored for HiSSD. The reported results are averaged over 5 random seeds. The bold number denotes the best performance, and $\pm$ denotes one standard deviation.}
\label{tab: hissdperf}
\begin{center}
\begin{small}
\begin{tabular}{cccc}
    \toprule
    $\bm\alpha$ & $\bm\epsilon$ & \textbf{Marine-Single} & \textbf{Marine-Single-Inv} \\
    \midrule
    $5.0$& $0.5$ & 38.25 $\pm$ 13.41  &  82.38 $\pm$ 4.28 \\
    $5.0$& $0.7$ & 43.50 $\pm$ 14.24  &  \textbf{86.88} $\pm$ 4.49 \\
    $5.0$& $0.9$ & 39.00 $\pm$ 14.09  &  82.38 $\pm$ 4.82 \\
    $10.0$& $0.5$ & 36.75 $\pm$ 7.78  &  83.25 $\pm$ 5.69 \\
    $10.0$& $0.7$ & 40.37 $\pm$ 15.25  &  74.63 $\pm$ 4.02 \\
    $10.0$& $0.9$ & \textbf{44.12} $\pm$ 18.89  & 85.25 $\pm$ 10.33	  \\
    $20.0$& $0.5$ & 39.88 $\pm$ 12.33  &  82.12 $\pm$ 3.53 \\
    $20.0$& $0.7$ & 37.00 $\pm$ 12.24  & 83.25 $\pm$ 5.13 \\
    $20.0$& $0.9$ & 37.50 $\pm$ 12.14  & 84.12 $\pm$ 7.67 \\
    \bottomrule
\end{tabular}
\end{small}
\end{center}
\end{table}

\newpage

\subsection{Computational Resources}\label{app:cr}
{We run SD-CQL on a single NVIDIA RTX A6000 GPU. Each run takes approximately 5 to 12 hours, depending on the task set (about 5 hours for the Marine-Easy-Expert task set and about 12 hours for the Stalker-Zealot-Medium-Expert task set).}   
\section{Skill-Discovery Visualization}\label{app:visual}
\subsection{Skill Visualization for SD-CQL}\label{app:sd_visual}
We visualize the {discovered skills} as follows: First, we deploy the SD-CQL agent, which has been trained on the \textit{Marine-Single} task set, in the \textit{3m} and \textit{12m} tasks. Then, we run one episode in each task and record the actions and skills of each agent at each time step, while saving replay videos. Next, we collect the skill $z$ of every agent at every time step from both trajectories as samples. Finally, we apply t-SNE \cite{tsne} to project them into a 2D plane and normalize the two dimensions to align their scales.

In Figure \ref{fig:z_visual}, the markers represent the skills of agents at corresponding time steps and tasks (Figure \ref{fig:3m_visual} and Figure \ref{fig:12m_visual}) after dimensionality reduction. Red markers represent skills from the \textit{12m} task, green markers represent skills from the \textit{3m} task, triangles indicate the \textit{\textbf{retreat}}, circles indicate the \textit{\textbf{attack}} skill, and crosses indicate dead agents. 

\begin{figure}[!thpb]
\begin{center}
    \includegraphics[width=.8\textwidth]{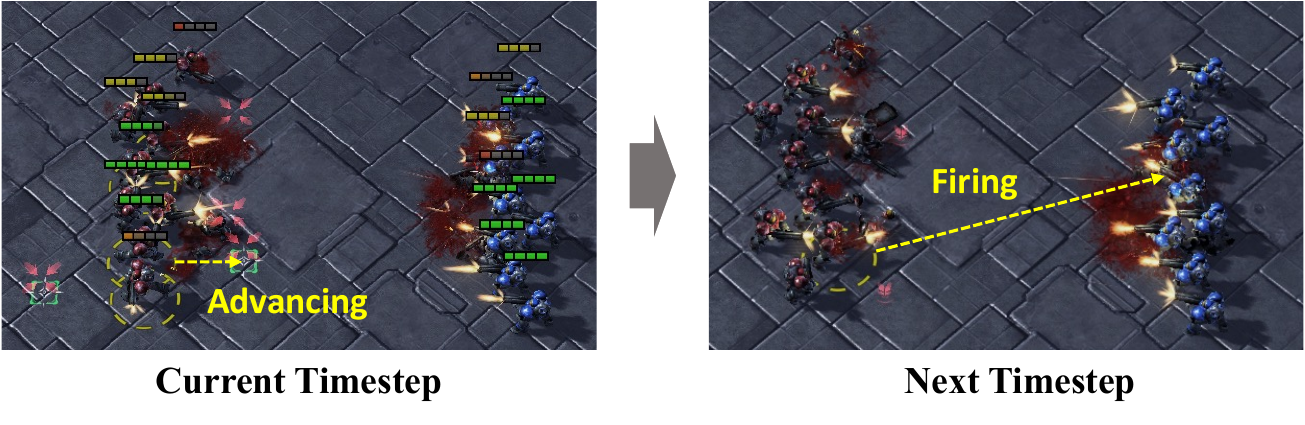}
    \caption{In the \textit{12m} task, the agents first approach the enemy through the \textit{advancing} action and then execute the \textit{firing} action in the next timestep.}
\label{fig:12m_evidence}
\end{center}
\end{figure}

{In Section \ref{exp_visual}, we mentioned that in the \textit{12m} task, after extracting the \textbf{\textit{attack}} skill, SD-CQL controls agents to perform two specific actions: \textit{firing} and \textit{advancing}. To validate this interpretation, we present in Figure \ref{fig:12m_evidence} the next-step actions of the marine units that have just executed the \textit{advancing} action. It can be observed that after moving forward to approach the enemies, they immediately perform a \textit{firing} action. This supports the interpretation in Figure \ref{fig:12m_visual} that the \textit{advancing} action is also part of the \textbf{\textit{attack}} skill. It also indicates that SD-CQL is capable of extracting a certain general skill (e.g., \textbf{\textit{attack}}) and flexibly executing different specific actions (e.g., \textit{firing} and \textit{advancing}) based on the situation.}

\subsection{Multi-timestep Visualization}
\begin{figure}[!thpb]
\begin{center}
    \begin{subfigure}[b]{\columnwidth}
        \centering
        \includegraphics[width=\columnwidth]{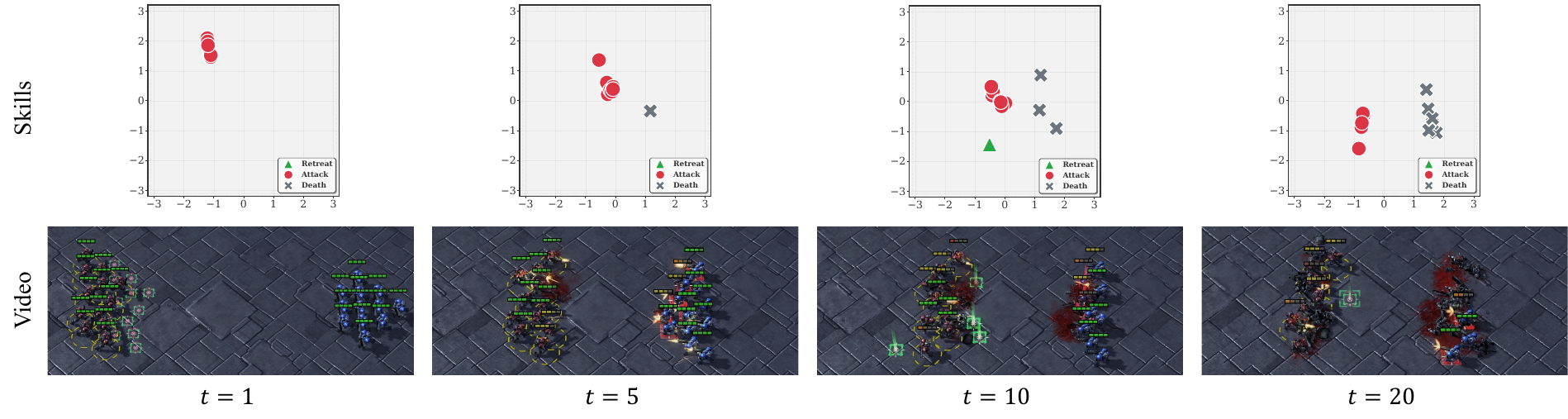}
        \caption{Visualization of SD-CQL skill representations and game replays across multiple timesteps.}\label{fig:sdcql_visual_comp}
    \end{subfigure} \\
    \vspace{.2cm}
    \begin{subfigure}[b]{\columnwidth}
        \centering
        \includegraphics[width=\columnwidth]{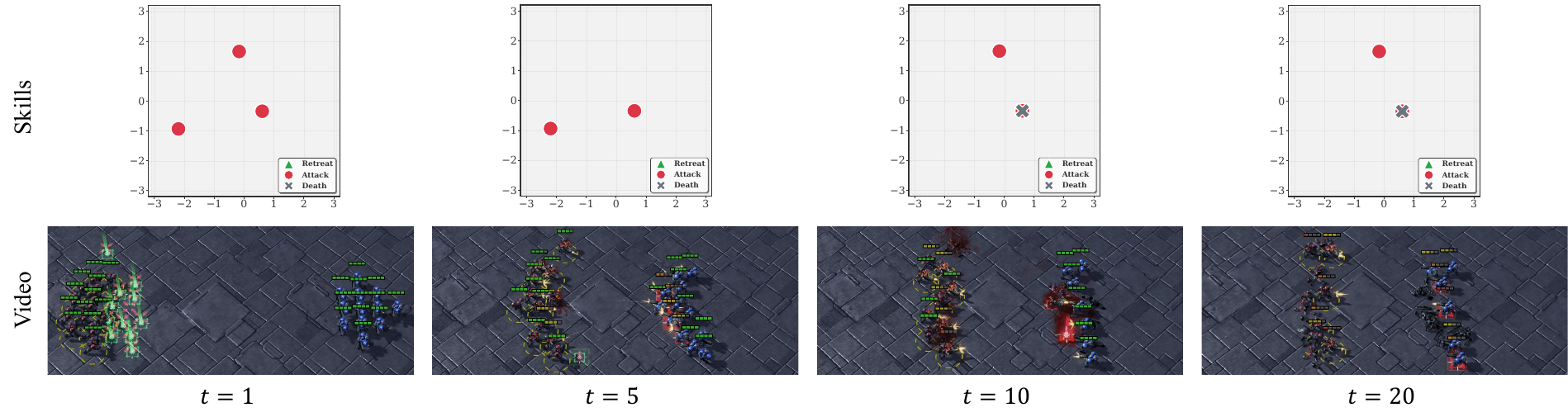}
        \caption{Visualization of ODIS skill representations and game replays across multiple timesteps.}\label{fig:odis_visual_comp}
    \end{subfigure}
    \caption{Visualization comparison of SD-CQL and ODIS across multiple timesteps in the \textit{12m} task.}
\label{fig:visual_comp}
\end{center}
\end{figure}

{To illustrate the consistency of SD-CQL’s skill-discovery mechanism across multiple timesteps, we present in Figure \ref{fig:sdcql_visual_comp} the game replays at different moments of the \textit{12m} task, alongside the 2D representations of the skills selected by each agent. Since HiSSD employs both common and task-specific skills, making a direct comparison with SD-CQL’s skills less straightforward, we instead visualize the skills and corresponding game replays of ODIS in Figure \ref{fig:odis_visual_comp} using a method similar to that described in Appendix \ref{app:sd_visual}.}

{By comparing the visualization results of SD-CQL and ODIS in Figure \ref{fig:visual_comp}, we observe that, due to the limited expressiveness of predefined discrete skills, ODIS fails to distinguish between dead and surviving agents. Moreover, we do not observe any agents performing retreat-like actions, which undermines the performance of ODIS. In contrast, the continuous skill vectors discovered by SD-CQL in the latent space offer much higher expressiveness, successfully differentiating agents in different states. Furthermore, by directly optimizing the skill-conditioned Q-values for specific actions, SD-CQL can flexibly execute different actions based on the situation after extracting certain skills, further enhancing its generalization performance and task efficiency.}

\section{Ablation Study}\label{app:ablation}
{To investigate the contribution of each component of SD-CQL to its superior performance, as well as its sensitivity to different hyperparameter values, we perform an ablation study on the One-to-Multi task sets.}

\subsection{Component Ablation}

\subsubsection{{Key Components of SD-CQL}}
In Table \ref{tab: abc-win-rates}, we report the performance of three variants of SD-CQL: (i) without the local value calibration term (w/o {LVC}), (ii) without the skill discovery mechanism (w/o SD), and (iii) without both (w/o {LVC} \& SD), {i.e., vanilla CQL}. It can be seen that these two components are indispensable for the superior performance of SD-CQL. {Across both task sets, our proposed skill discovery mechanism and corresponding skill-conditioned Q-learning significantly enhance the generalization ability of SD-CQL. Meanwhile, the {LVC} effectively mitigates error accumulation, particularly in the \textit{Marine-Single-Inv} task set, where the source task involves a large number of agents.}

{In addition, we observe some interesting phenomena. For example, the w/o SD \& {LVC} variant achieves the second-best performance after SD-CQL on the \textit{Marine-Single} task set, and it also outperforms the w/o {LVC} variant on the \textit{Marine-Single-Inv} task set. These results may be attributed to differences in the expressiveness of the model.}

{Specifically, as shown in Figure \ref{fig:sdcql}, in w/o {LVC}, the encoder is updated only by $\mathcal{L}_{\text{Rec}}$ according to Eq. \eqref{rec_loss}, while Q-values are optimized by $\mathcal{L}_{\text{Q}}$ according to Eq. \eqref{q_learning}. However, in w/o SD \& {LVC}, the removal of $\mathcal{L}_{\text{Rec}}$ necessitates eliminating the gradient stop before Q-values shown in Figure \ref{fig:sdcql}, enabling the encoder and Q-values to be jointly optimized. As a result, the policy network possesses more trainable parameters, providing greater expressiveness. This explains why w/o SD \& {LVC} outperforms w/o {LVC}. {Overall}, SD-CQL introduces {local value calibration} to bridge skill discovery and Q-learning, thereby enhancing overall cross-task performance and highlighting our key innovation.}

\begin{table}[!htpb]
\caption{Win rates of different SD-CQL variants on One-to-Multi task sets. The reported results are averaged over 5 random seeds. The bold number denotes the best performance, and $\pm$ denotes one standard deviation.}
\label{tab: abc-win-rates}
\vskip 0.1in
\begin{center}
\begin{small}
\begin{tabular}{clccccc}
\toprule
\multicolumn{2}{c}{\textbf{Task Set}} & \textbf{SD-CQL} & \textbf{w/o {LVC}} & \textbf{w/o SD} & \textbf{w/o {LVC} \& SD} \\
\midrule
\multirow{5}{*}{\rotatebox{90}{\makecell{\textit{Marine}\\\textit{Single}}}} 
    &3m $^\diamond$ & \textbf{100.00} $\pm$ 0.00 & 98.12 $\pm$ 2.80 & 99.38 $\pm$ 1.40 & \textbf{100.00} $\pm$ 0.00 \\
    &5m & \textbf{91.25} $\pm$ 10.92 & 83.75 $\pm$ 17.17 & 60.62 $\pm$ 41.79 & 88.12 $\pm$ 7.46 \\
    &8m & \textbf{49.38} $\pm$ 23.63 & 39.38 $\pm$ 17.34 & 33.75 $\pm$ 30.49 & 40.00 $\pm$ 26.00\\
    &10m & \textbf{41.88} $\pm$ 25.83 & 27.50 $\pm$ 18.41 & 31.87 $\pm$ 36.54 & 34.38 $\pm$ 25.39 \\
    &12m  & \textbf{27.50} $\pm$ 21.13 & 17.50 $\pm$ 14.76 & 10.00 $\pm$ 9.73 & 17.50 $\pm$ 10.27 \\
    \midrule
    \multicolumn{2}{c}{\textbf{Average}}  & \textbf{62.00} $\pm$ 13.80 & 53.25 $\pm$ 11.47 & 47.12 $\pm$ 20.16 & 56.00 $\pm$ 11.07 \\
\midrule\midrule
\multirow{5}{*}{\rotatebox{90}{\makecell{\textit{Marine} \\ \textit{Single-Inv}}}}
    &10m $^\diamond$ &\textbf{100.00} $\pm$ 0.00 & 0.00 $\pm$ 0.00 & \textbf{100.00} $\pm$ 0.00& 0.62 $\pm$ 1.4 \\
    &8m  &\textbf{100.00} $\pm$ 0.00 & 0.00 $\pm$ 0.00 & 98.75 $\pm$ 2.8 & 3.75 $\pm$ 8.39 \\
    &5m  &\textbf{97.50} $\pm$ 2.61 & 0.00 $\pm$ 0.00 & 26.25 $\pm$ 20.32 & 6.25 $\pm$ 13.98 \\
    &4m  &\textbf{75.62} $\pm$ 21.58 & 0.00 $\pm$ 0.00 & 3.12 $\pm$ 6.99 & 0.62 $\pm$ 1.4 \\
    &3m  &\textbf{64.38} $\pm$ 15.40 & 0.00 $\pm$ 0.00 & 0.0 $\pm$ 0.0 & 2.5 $\pm$ 4.07 \\
    \midrule
    \multicolumn{2}{c}{\textbf{Average}}  &\textbf{87.50} $\pm$ 6.54 & 0.00 $\pm$ 0.00 & 45.62 $\pm$ 4.73& 2.75 $\pm$ 5.19\\
\bottomrule
\multicolumn{5}{l}{\small $\diamond$ denotes the source task.}\\
\end{tabular}
\end{small}
\end{center}
\vskip -0.1in
\end{table}

\subsubsection{{Design Choices in Reconstructor}}
{SD-CQL extracts effective skills by reconstructing the next local observation. To investigate the role of several design choices in the reconstructor, such as the reconstruction target $o_{t+1}$, the temporal latent variable $h$, and the skill vector $z$, we conduct additional ablation studies. Specifically, we evaluate four variants on the marine-single and marine-single-inv task sets: (i) Reconstructing the current observation instead of the next (Rec $o_t$); (ii) Removing the temporal latent variable (w/o $h$); (iii) Not feeding the skill vector into the Q-network (No $z$ to Q); and (iv) Performing reconstruction only without extracting any skill vector (Only Rec).}

{Apart from the modifications specific to each variant, all other components, such as network architecture and hyperparameters, remain the same as in SD-CQL. We report the average performance of these variants across both task sets in Table \ref{tab: abrec}.}

{It can be observed that the complete SD-CQL achieves the highest overall performance, and skill extraction is critical to its superior performance: all variants that incorporate the skill vector into the Q-function outperform those that do not. Even when the skill vector is excluded from the Q-function, incorporating it into the reconstruction process still helps extract more informative individual features. At the same time, predicting the next local observation and incorporating temporal latent variables indeed help SD-CQL better capture generalizable decision-making patterns.}
\begin{table}[!htpb]
\caption{The average win rates of different SD-CQL variants on One-to-Multi task sets. The reported results are averaged over 5 random seeds. The bold number denotes the best performance, and $\pm$ denotes one standard deviation.}
\label{tab: abrec}
\vskip 0.1in
\begin{center}
\begin{small}
\begin{tabular}{clcccccc}
\toprule
\multicolumn{2}{c}{\textbf{Task Set}} & \textbf{SD-CQL} & \textbf{Rec $o_t$} & \textbf{w/o $h$} & \textbf{No $z$ to Q} & \textbf{Only Rec} \\
\midrule
\multicolumn{2}{c}{\textit{Marine-Single}}  & 62.00 $\pm$ 13.80 & 54.88 $\pm$ 15.27 & \textbf{62.25} $\pm$ 16.70 & 62.13 $\pm$ 15.17& 40.88 $\pm$ 15.17\\
\midrule
\multicolumn{2}{c}{\textit{Marine-Single-Inv}}  &\textbf{87.50} $\pm$ 6.54 & 87.38 $\pm$ 4.06& 81.38 $\pm$ 8.69& 76.25 $\pm$ 9.60& 64.63 $\pm$ 23.69\\
\midrule\midrule
\multicolumn{2}{c}{\textbf{Average}}& \textbf{74.75} $\pm$ 6.57 & 71.13 $\pm$ 8.87 & 71.94 $\pm$ 4.89& 69.19 $\pm$ 9.77& 52.75 $\pm$ 14.88\\
\bottomrule
\end{tabular}
\end{small}
\end{center}
\vskip -0.1in
\end{table}

\subsection{Hyperparameter Sensitivity}
In Table \ref{tab: abh-win-rates}, we separately adjusted one of the weights of the CQL regularization $\alpha$ and the {LVC} $\eta$ while keeping the other consistent with that in the evaluation. {It can be observed that the performance of SD-CQL remains relatively stable and insensitive to the values of these two hyperparameters on the informative \textit{Marine-Single-Inv} task set. In contrast, on the more limited \textit{Marine-Single} task set, hyperparameter tuning requires slightly more care. However, it is not particularly challenging, as it is reasonable to adopt a fairly conservative $\alpha$ on expert datasets and apply less {LVC} regularization $\eta$ when dealing with a small number of agents, given the lower level of accumulated error compared to \textit{Marine-Single-Inv}.}
				
\begin{table}[!htbp]
\caption{Win rates of SD-CQL with different hyperparameters on One-to-Multi task sets. The reported results are averaged over 5 random seeds. The $\pm$ denotes one standard deviation.}
\label{tab: abh-win-rates}
\vskip 0.1in
\begin{center}
\begin{small}
\begin{tabular}{llcccc}
    \toprule
    \multicolumn{2}{c}{\textbf{Hyperparameters}} & $\eta=0.7$ & $\eta=0.5$ & $\eta=0.3$\\
    \midrule
    \multirow{2}{*}{\textbf{Average}} & \textit{Marine-Single}& 45.50 $\pm$ 9.32 & 62.00 $\pm$ 13.8 & 62.50 $\pm$ 16.28 \\
     & \textit{Marine-Single-Inv}& 83.50 $\pm$ 8.67 & 84.00 $\pm$ 9.64 & 87.50 $\pm$ 6.54\\
    \midrule
    \midrule
    \multicolumn{2}{c}{\textbf{Hyperparameters}} & $\alpha=0.5$ & $\alpha=1.0$ & $\alpha=2.5$\\
    \midrule
    \multirow{2}{*}{\textbf{Average}} & \textit{Marine-Single}&  51.88 $\pm$ 12.52 & 62.00 $\pm$ 13.8 & 56.12 $\pm$ 13.45 \\
    & \textit{Marine-Single-Inv}& 81.12 $\pm$ 4.43 & 87.50 $\pm$ 6.54 & 78.88 $\pm$ 10.69\\
    \bottomrule
\end{tabular}
\end{small}
\end{center}
\vskip -0.1in
\end{table}

\section{Discussions}\label{app:dis}
\subsection{{Assumptions and Scope}}
{We consider the standard multi-task MARL setting, in line with prior work \cite{hu2021updet,zhang2023odis,chen2024variational,hissd}, where tasks, whether in online or offline regimes, share a similar and well-defined observation structure that enables cross-task generalization of policy models. When this assumption is violated, such as in less structured, vision-based settings, ambiguous or incorrect entity decomposition may hinder performance for existing approaches, including SD-CQL. This challenge is shared across current multi-task MARL methods. Nevertheless, techniques such as entity segmentation and representation learning can transform noisy or unstructured inputs into structured representations that SD-CQL can exploit to learn generalizable policies. We leave this as an important direction for future work.}

{Even under well-defined observations, multi-task offline MARL remains highly challenging and largely open. Our experiments show that existing methods often fall short in this regime, while SD-CQL delivers substantial gains and establishes a new state of the art.}

\subsection{Detailed Architecture Comparison with ODIS, HiSSD, and VO-MASD}\label{app:comp}
\subsubsection{Comparison with ODIS}
{ODIS requires a pre-specified number of skills and learns a skill classifier and action decoder by reconstructing actions from global states. During policy training, it freezes the classifier, retrains a local encoder, and uses CQL to choose skills, decoding actions through the decoder.}

{In contrast, SD-CQL directly discovers skills in the latent space via a local reconstructor. It then chooses actions using skill-conditioned CQL with {local value calibration}. This eliminates the need for predefined skill numbers or retraining, and achieves better generalization and task efficiency.}
\subsubsection{Comparison with HiSSD}
{HiSSD comprises a high-level planner and a low-level controller. The high-level planner predicts the distribution of the next global state from the current observation and trains an IQL-style value function to weight the likelihood loss, thereby extracting common skills. The low-level controller learns task-specific skills via contrastive learning and generates actions through a BC decoder conditioned on the observation and both types of skills.}

{Although HiSSD also operates in a latent skill space, it exhibits several limitations compared to SD-CQL. First, like ODIS, it relies on BC imitation for action generation, limiting its ability to exceed the behavior policy. Second, its contrastive learning mechanism cannot be applied when only a single source task is available. Finally, its complex architecture inflates the parameter count to over four times that of ODIS and SD-CQL, leading to longer training times and lower efficiency.}

{In contrast, SD-CQL employs a simpler yet more effective design: it discovers skills through observation reconstruction and directly optimizes skill-conditioned Q-values with {local value calibration}, achieving better generalization and task efficiency with significantly fewer parameters and shorter training durations.}

\subsubsection{{Comparison with VO-MASD}}
{VO-MASD targets multi-task MARL by discovering offline skills via an improved VQ-VAE and then training target-task high-level policies online with a hierarchical PPO. This is fundamentally different from our multi-task offline setting, where learning is completed without any online interaction and policies are deployed zero-shot.}

{Beyond the training protocol, the core technical designs also diverge. VO-MASD reconstructs actions from a predefined number of codebooks, implicitly assuming a fixed discrete skill inventory.}

{In contrast, SD-CQL reconstructs the next-step observation in a continuous latent space, enabling flexible and scalable skill discovery without fixing the number of skills, and then optimizes skill-conditioned policies entirely offline. Our results show that, when coupled with next-observation–based offline skill discovery and regularized by local value calibration,} {value-based method} {is not only viable but superior in multi-task offline MARL.}

{Nevertheless, exploring alternative skill-discovery modules within SD-CQL, as well as hierarchical offline-to-online extensions, is a promising direction for future work.}

\subsection{``Task label'' in (Offline) Multi-Task MARL and Conventional RL}\label{app:multi}

{The ``task label'' is an essential concept in multi-task RL, as it explicitly indicates the task an agent is solving and guides policy learning across diverse objectives. Although multi-task MARL (MT-MARL) typically lacks such explicit task labels, the underlying tasks can still be related to those in conventional multi-task RL by abstracting them at two levels. The first level focuses on decision patterns based on the current team state and executing corresponding actions within a specific task, and the second level corresponds to team-level objectives that differ across tasks. The core objective of MT-MARL is to discover and leverage the decision patterns that generalize across tasks to enable multi-task generalization and task efficiency.}

{Therefore, the concept of ``task label'' in MT-MARL may share some relationship with these first-level decision patterns, i.e., ``skills''. As illustrated in Figure \ref{fig:moti}, the agents’ behaviors naturally cluster into distinct skills, and they must infer the appropriate skill from the current observation and accurately execute the corresponding actions, which is fundamentally similar to conventional single-agent multi-task RL, where different task objectives require different action policies.}

{However, note that the ``skills'' are not exactly equivalent to ``task labels'' in the conventional sense. Since task labels in conventional multi-task RL settings are typically discrete, pre-defined, and well-specified, whereas in MT-MARL, the skill boundaries are usually not explicitly available. As shown in Figures \ref{fig:single} and \ref{fig:visual_comp}, prior methods that predefine the number of skills and discover them through classification struggle to accurately capture the characteristics of each skill, thereby hindering generalization. In contrast, leveraging continuous representations in the latent space enables more effective identification of these decision patterns and improves generalization. This also highlights that the design of SD-CQL is tailored to address the unique challenges of MT-MARL.}

\newpage
\section{Detailed Results of Multi-to-Multi Task Sets}\label{app:results}
\subsection{Multi-to-Multi Task Sets}\label{app:results-mtm}
We present the detailed results of our evaluation experiments on the Multi-to-Multi task sets. The results for \textit{Marine-Easy} are presented in Table \ref{tab:e-win-rates}, the results for \textit{Marine-Hard} are shown in Table \ref{tab:h-win-rates}, and the results for \textit{Stalker-Zealot} are provided in Table \ref{tab:sz-win-rates}. In each table, we present the multi-task evaluation results on datasets of varying quality, with the source tasks used for training marked by ``\(\diamond\)''. The results show that, despite being trained on a limited number of source tasks, SD-CQL exhibits strong multi-task generalization, as reflected in its exceptional average performance across all test tasks.
\begin{table*}[htbp!]
\caption{Win rates of \textit{Marine-Easy} {scenario}. The reported results are averaged over 5 random seeds. The bold number denotes the best performance, and $\pm$ denotes one standard deviation.}
\label{tab:e-win-rates}
\vskip 0.15in
\begin{center}
\resizebox{.9\linewidth}{!}{%
\begin{tabular}{clcccccc}
    \toprule
    \multicolumn{2}{c}{\textbf{Task}} & \textbf{BC-t} & \textbf{BC-r}  &\textbf{CQL}&\textbf{ODIS} & \textbf{HiSSD}& \textbf{SD-CQL (Ours)} \\
    \midrule
    \multirow{10}{*}{\rotatebox{90}{\makecell{Expert}}}
    &3m $^\diamond$& 99.38 $\pm$ 1.40 & \textbf{100.00} $\pm$ 0.00  &98.13 $\pm$ 2.80& 97.50 $\pm$ 1.40 & 98.75 $\pm$ 2.80& 98.75 $\pm$ 2.80
\\  
    &4m & 90.62 $\pm$ 7.33 & 96.88 $\pm$ 3.83  &80.63 $\pm$ 13.69& 56.25 $\pm$ 37.69 & 92.5 $\pm$ 8.73& \textbf{97.50} $\pm$ 2.61
\\
    &5m $^\diamond$& \textbf{100.00} $\pm$ 0.00 & \textbf{100.00} $\pm$ 0.00  &81.25 $\pm$ 11.48& 78.12 $\pm$ 31.48 & \textbf{100.00} $\pm$ 0.00& 98.75 $\pm$ 1.71
\\
    &6m & \textbf{100.00} $\pm$ 0.00 & \textbf{100.00} $\pm$ 0.00  &40.00 $\pm$ 30.89& 55.62 $\pm$ 43.49 & 98.75 $\pm$ 1.71& 96.88 $\pm$ 4.42
\\
    &7m & \textbf{100.00} $\pm$ 0.00 & 98.75 $\pm$ 1.71  &10.63 $\pm$ 23.76& 58.12 $\pm$ 44.28 & 98.75 $\pm$ 2.80& 97.50 $\pm$ 4.07
\\
    &8m & 99.38 $\pm$ 1.40 & \textbf{100.00} $\pm$ 0.00  &0.63 $\pm$
1.40 & 63.75 $\pm$ 41.14 & \textbf{100.00} $\pm$ 0.00& 98.12 $\pm$ 1.71
\\
    &9m & \textbf{100.00} $\pm$ 0.00 & \textbf{100.00} $\pm$ 0.00  &0.00 $\pm$ 0.00& 70.62 $\pm$ 42.65 & 99.38 $\pm$ 1.40& \textbf{100.00} $\pm$ 0.00
\\
    &10m $^\diamond$& \textbf{100.00} $\pm$ 0.00 & 98.75 $\pm$ 1.71  &0.00 $\pm$ 0.00& 78.75 $\pm$ 35.24 & 98.75 $\pm$ 1.71& 98.12 $\pm$ 2.80
\\
    &11m & 99.38 $\pm$ 1.40 & \textbf{100.00} $\pm$ 0.00  &0.00 $\pm$ 0.00& 80.00 $\pm$ 28.78 & 98.75 $\pm$ 1.71& 98.12 $\pm$ 4.19
\\
    &12m & \textbf{100.00} $\pm$ 0.00 & 99.38 $\pm$ 1.40  &0.00 $\pm$ 0.00& 66.25 $\pm$ 42.01 & 93.75 $\pm$ 9.11& 96.88 $\pm$ 2.21
\\
    \midrule
\multicolumn{2}{c}{\textbf{Average}} & 98.87 $\pm$ 0.58 & \textbf{99.38} $\pm$ 0.40  &31.13 $\pm$ 6.36& 70.50 $\pm$ 30.14 & 97.94 $\pm$ 1.23& 98.06 $\pm$ 1.00\\
    \midrule
    \multirow{10}{*}{\rotatebox{90}{\makecell{Medium}}}
    &3m$^\diamond$& 61.25 $\pm$ 10.03 & 65.62 $\pm$ 8.84  &\textbf{70.63} $\pm$ 17.48& 64.38 $\pm$ 10.96 & 61.88 $\pm$ 5.59& 70.62 $\pm$ 12.62\\
    &4m & \textbf{66.25} $\pm$ 8.09 & 64.38 $\pm$ 34.42  &27.50 $\pm$ 30.09& 55.00 $\pm$ 30.98 & 71.88 $\pm$ 10.13& 52.50 $\pm$ 33.18
\\
    &5m $^\diamond$ & 78.75 $\pm$ 8.09 & 80.00 $\pm$ 4.19  &43.75 $\pm$ 34.59& 76.25 $\pm$ 7.19 & 83.12 $\pm$ 9.78& \textbf{84.38} $\pm$ 4.94
\\
    &6m & 90.00 $\pm$ 6.01 & 84.38 $\pm$ 5.85  &24.38 $\pm$ 23.43& \textbf{93.12} $\pm$ 8.39 & 76.88 $\pm$ 12.02& 89.38 $\pm$ 9.27
\\
    &7m    & 99.38 $\pm$ 1.40 & 90.62 $\pm$ 11.69  &7.50 $\pm$ 9.00& 90.62 $\pm$ 12.50 & 95.00 $\pm$ 6.48& \textbf{100.00} $\pm$ 0.00
\\
    &8m    & 86.88 $\pm$ 10.69 & \textbf{95.62} $\pm$ 6.48  &0.00 $\pm$ 0.00& 89.38 $\pm$ 5.23 & 93.12 $\pm$ 4.07& 90.00 $\pm$ 6.40
\\
    &9m     & \textbf{83.12} $\pm$ 8.73 & 78.75 $\pm$ 5.13  &0.00 $\pm$ 0.00& 72.50 $\pm$ 18.67 & 76.88 $\pm$ 7.84& 82.50 $\pm$ 13.00
\\
    &10m $^\diamond$& 70.62 $\pm$ 17.34 & 69.38 $\pm$ 6.77  &0.00 $\pm$ 0.00& 70.00 $\pm$ 15.08 & 75.62 $\pm$ 6.01& \textbf{87.50} $\pm$ 6.25
\\
    &11m    & 43.75 $\pm$ 13.07 & 43.75 $\pm$ 5.85  &0.00 $\pm$ 0.00& 43.12 $\pm$ 9.73 & 46.25 $\pm$ 12.18& \textbf{55.00} $\pm$ 10.27
\\
    &12m   & 38.12 $\pm$ 7.78 & 44.38 $\pm$ 5.59  &0.00 $\pm$ 0.00& 30.62 $\pm$ 12.38 & 33.75 $\pm$ 5.59& \textbf{46.25} $\pm$ 10.69
\\
    \midrule
\multicolumn{2}{c}{\textbf{Average}} & 71.81 $\pm$ 3.74 & 71.69 $\pm$ 4.33  &17.38 $\pm$ 7.31& 68.50 $\pm$ 5.74 & 71.44 $\pm$ 2.77& \textbf{75.81} $\pm$ 1.73\\
 
    \midrule
    \multirow{10}{*}{\rotatebox{90}{\makecell{Medium-Replay}}}
    &3m $^\diamond$& 80.00 $\pm$ 6.48 & 70.62 $\pm$ 22.81  &71.88 $\pm$ 11.05& 60.62 $\pm$ 35.05 & \textbf{82.50} $\pm$ 9.53& 78.75 $\pm$ 16.45
\\
    &4m & 75.00 $\pm$ 15.15 & 70.00 $\pm$ 9.53  &74.38 $\pm$ 9.48& 18.75 $\pm$ 25.77 & 69.38 $\pm$ 21.13& \textbf{79.38} $\pm$ 7.53
\\
    &5m $^\diamond$& 77.50 $\pm$ 31.51 & 88.75 $\pm$ 15.56  &62.50 $\pm$ 14.82& 73.75 $\pm$ 26.01 & 86.25 $\pm$ 7.19& \textbf{91.25} $\pm$ 9.73
\\
    &6m & 88.75 $\pm$ 23.45 & 98.12 $\pm$ 2.80  &40.63 $\pm$ 41.93& 12.50 $\pm$ 27.95 & \textbf{100.00} $\pm$ 0.00& 99.38 $\pm$ 1.40
\\
    &7m & 86.88 $\pm$ 24.15 & 90.62 $\pm$ 12.88  &26.25 $\pm$ 36.01& 8.12 $\pm$ 18.17 & \textbf{100.00} $\pm$ 0.00& \textbf{100.00} $\pm$ 0.00
\\
    &8m & 19.38 $\pm$ 29.43 & 43.75 $\pm$ 35.84  &9.38 $\pm$ 20.96& 0.00 $\pm$ 0.00 & \textbf{96.25} $\pm$ 5.13& 89.38 $\pm$ 10.27
\\
    &9m & 15.62 $\pm$ 18.88 & 25.00 $\pm$ 24.51  &3.75 $\pm$ 8.39& 5.00 $\pm$ 11.18 & 76.25 $\pm$ 4.74& \textbf{88.75} $\pm$ 4.19
\\
    &10m $^\diamond$& \textbf{80.62} $\pm$ 12.18 & 76.88 $\pm$ 17.62  &3.13 $\pm$ 6.99& 76.88 $\pm$ 11.82 & 68.12 $\pm$ 10.92& 70.00 $\pm$ 39.26
\\
    &11m & 1.25 $\pm$ 2.80 & 5.00 $\pm$ 8.15  &3.13 $\pm$ 6.99& 0.00 $\pm$ 0.00 & 24.38 $\pm$ 13.15& \textbf{35.62} $\pm$ 19.96
\\
    &12m & 0.00 $\pm$ 0.00 & 3.75 $\pm$ 6.77  &0.63 $\pm$ 1.40& 0.00 $\pm$ 0.00 & \textbf{25.62} $\pm$ 23.74& \textbf{25.62} $\pm$ 15.05
\\
    \midrule
    \multicolumn{2}{c}{\textbf{Average}} & 44.12 $\pm$ 8.19 & 49.62 $\pm$ 9.17  &29.56 $\pm$ 11.73& 11.06 $\pm$ 8.94 & 72.88 $\pm$ 3.16& \textbf{75.81} $\pm$ 6.83\\
    \midrule
    \multirow{10}{*}{\rotatebox{90}{\makecell{Medium-Expert}}}&3m $^\diamond$ & 90.00 $\pm$ 18.93 & 83.75 $\pm$ 23.94  &\textbf{96.88} $\pm$ 3.13& 63.12 $\pm$ 27.90 & 90.62 $\pm$ 20.96& \textbf{96.88} $\pm$ 3.12
\\
    &4m & \textbf{87.50} $\pm$ 9.11 & 78.12 $\pm$ 14.82  &55.63 $\pm$ 24.55& 49.38 $\pm$ 37.72 & 76.25 $\pm$ 12.62& 72.50 $\pm$ 19.19
\\
    &5m $^\diamond$& 69.38 $\pm$ 9.22 & 77.50 $\pm$ 10.69  &70.00 $\pm$ 29.03& 5.00 $\pm$ 11.18 & 88.12 $\pm$ 14.89& \textbf{96.25} $\pm$ 6.77
\\
    &6m & 63.75 $\pm$ 25.92 & 68.75 $\pm$ 20.73  &80.00 $\pm$ 15.08& 50.00 $\pm$ 9.88 & 55.00 $\pm$ 15.08& \textbf{93.75} $\pm$ 6.25
\\
    &7m & 85.62 $\pm$ 6.85 & 86.25 $\pm$ 16.33  &31.25 $\pm$ 17.26& 65.62 $\pm$ 22.53 & 86.25 $\pm$ 7.84& \textbf{90.62} $\pm$ 11.05
\\
    &8m & 56.88 $\pm$ 11.14 & 66.25 $\pm$ 24.65  &1.88 $\pm$ 4.19& 66.25 $\pm$ 23.84 & 76.25 $\pm$ 8.15& \textbf{90.00} $\pm$ 9.73
\\
    &9m & 70.62 $\pm$ 12.62 & 75.00 $\pm$ 22.43  &0.00 $\pm$ 0.00& 73.12 $\pm$ 18.43 & \textbf{90.62} $\pm$ 5.85& 87.50 $\pm$ 10.60
\\
    &10m $^\diamond$& 5.00 $\pm$ 6.85 & 11.88 $\pm$ 12.96  &0.00 $\pm$ 0.00& 0.62 $\pm$ 1.40 & 84.38 $\pm$ 16.83& \textbf{87.50} $\pm$ 11.90
\\
    &11m & 68.12 $\pm$ 14.22 & 71.88 $\pm$ 8.56  &0.00 $\pm$ 0.00& 76.88 $\pm$ 18.83 & 75.00 $\pm$ 19.52& \textbf{81.88} $\pm$ 17.73
\\
    &12m & 60.62 $\pm$ 12.81 & 56.88 $\pm$ 4.64  &0.00 $\pm$ 0.00& 65.00 $\pm$ 16.00 & 60.62 $\pm$ 12.81& \textbf{71.88} $\pm$ 18.09
\\
    \midrule
\multicolumn{2}{c}{\textbf{Average}}  & 74.12 $\pm$ 3.21 & 75.25 $\pm$ 7.41  &33.56 $\pm$ 6.47& 66.00 $\pm$ 9.21 & 78.31 $\pm$ 4.99& \textbf{86.88} $\pm$ 5.68\\
    \bottomrule
    \multicolumn{7}{l}{\small $\diamond$ denotes the source tasks.} &
    \end{tabular}
}
\end{center}
\vskip -0.1in
\end{table*}

\begin{table}[htbp!]
\caption{Win rates of \textit{Marine-Hard} {scenario}. The reported results are averaged over 5 random seeds. The bold number denotes the best performance, and $\pm$ denotes one standard deviation.}
\label{tab:h-win-rates}
\vskip 0.1in
\begin{center}
\resizebox{.9\linewidth}{!}{%
\begin{tabular}{clcccccc}
    \toprule
    \multicolumn{2}{c}{\textbf{Task}} & \textbf{BC-t} & \textbf{BC-r}  &\textbf{CQL}&\textbf{ODIS} & \textbf{HiSSD}& \textbf{SD-CQL (Ours)} \\
    \midrule
    \multirow{12}{*}{\rotatebox{90}{\makecell{Expert}}}
    &3m $^\diamond$& \textbf{100.00} $\pm$ 0.00 & \textbf{100.00} $\pm$ 0.00  &98.12 $\pm$ 2.8& 74.38 $\pm$ 34.47 & 98.12 $\pm$ 1.71
& 99.38 $\pm$ 1.40
\\
    &4m & 92.50 $\pm$ 8.44 & 95.00 $\pm$ 7.19  &92.50 $\pm$ 7.19& 37.50 $\pm$ 32.40 & 98.75 $\pm$ 1.71
& \textbf{99.38} $\pm$ 1.40
\\
    &5m & 97.50 $\pm$ 2.61 & 83.75 $\pm$ 12.38  &96.25 $\pm$ 5.13& 30.62 $\pm$ 32.73 & \textbf{100.0} $\pm$ 0.00
& 99.38 $\pm$ 1.40
\\
    &10m & 95.62 $\pm$ 3.56 & 91.25 $\pm$ 13.15  &7.50 $\pm$ 13.55& 18.12 $\pm$ 12.96 & \textbf{99.38} $\pm$ 1.40
& 95.00 $\pm$ 9.53
\\
    &12m & 68.75 $\pm$ 31.25 & \textbf{88.75} $\pm$ 10.73  &1.88 $\pm$ 4.19& 5.00 $\pm$ 11.18 & 53.75 $\pm$ 50.7
& 74.38 $\pm$ 38.04
\\
    &5m6m $^\diamond$& 62.50 $\pm$ 8.84 & 67.50 $\pm$ 8.44  &0.62 $\pm$ 1.40& 25.00 $\pm$ 29.06 & \textbf{79.38} $\pm$ 6.48
& 60.00 $\pm$ 9.73
\\
    &7m8m & 30.62 $\pm$ 17.73 & 23.12 $\pm$ 19.84  &0.62 $\pm$ 1.40& 8.75 $\pm$ 11.14 & \textbf{37.50} $\pm$ 9.38
& \textbf{37.50} $\pm$ 16.24
\\
    &8m9m & 41.88 $\pm$ 6.85 & 43.75 $\pm$ 19.01  &0.00 $\pm$ 0.00& 15.00 $\pm$ 15.84 & 53.12 $\pm$ 16.09
& \textbf{87.50} $\pm$ 14.32
\\
    &9m10m $^\diamond$& \textbf{98.75} $\pm$ 2.80 & 90.00 $\pm$ 12.38  &0.62 $\pm$ 1.40& 23.75 $\pm$ 36.55 & \textbf{98.75} $\pm$ 1.71
& 95.00 $\pm$ 1.71
\\
    &10m11m & 63.75 $\pm$ 7.84 & \textbf{86.25} $\pm$ 8.44  &0.00 $\pm$ 0.00& 12.50 $\pm$ 14.66 & 83.75 $\pm$ 16.89
& 75.62 $\pm$ 9.73
\\
    &10m12m & 3.75 $\pm$ 2.61 & 8.12 $\pm$ 5.68  &0.00 $\pm$ 0.00& 1.88 $\pm$ 4.19 & 10.00 $\pm$ 4.64
& \textbf{23.12} $\pm$ 16.77
\\
    &13m15m & 0.00 $\pm$ 0.00 & 0.62 $\pm$ 1.40  &0.00 $\pm$ 0.00& 0.62 $\pm$ 1.40 & \textbf{3.75} $\pm$ 5.13
& 1.88 $\pm$ 2.80
\\
    \midrule
\multicolumn{2}{c}{\textbf{Average}} & 62.97 $\pm$ 2.18 & 64.84 $\pm$ 3.24  &24.84 $\pm$ 1.53& 21.09 $\pm$ 14.72 & 68.02 $\pm$ 4.47& \textbf{70.68} $\pm$ 4.80\\
    \midrule
    \multirow{12}{*}{\rotatebox{90}{\makecell{Medium}}} 
    &3m $^\diamond$& 62.50 $\pm$ 9.38 & 52.50 $\pm$ 15.21  &\textbf{81.25} $\pm$ 6.63& 59.38 $\pm$ 7.97 & 64.38 $\pm$ 5.23
& 75.62 $\pm$ 8.95\\
     &4m & 41.25 $\pm$ 23.32 & 41.88 $\pm$ 25.92  &8.12 $\pm$ 4.74& 58.12 $\pm$ 24.37 & \textbf{76.88} $\pm$ 7.53
& 71.88 $\pm$ 7.97
\\
    &5m & 86.88 $\pm$ 6.01 & 85.62 $\pm$ 11.40  &18.75 $\pm$ 17.26& 68.12 $\pm$ 30.49 & 90.62 $\pm$ 12.69
& \textbf{96.25} $\pm$ 5.59
\\
    &10m & 90.00 $\pm$ 12.77 & 90.00 $\pm$ 12.38  &0.00 $\pm$ 0.00& 70.00 $\pm$ 32.67 & 93.75 $\pm$ 3.12
& \textbf{96.25} $\pm$ 8.39
\\
    &12m & 75.00 $\pm$ 21.31 & 64.38 $\pm$ 30.51  &1.88 $\pm$ 2.80& 18.12 $\pm$ 23.74 & \textbf{79.38} $\pm$ 9.53
& 72.50 $\pm$ 21.81
\\
    &5m6m $^\diamond$& 30.62 $\pm$ 2.61 & \textbf{38.75} $\pm$ 6.48  &0.00 $\pm$ 0.00& 10.00 $\pm$ 11.98 & 30.63 $\pm$ 11.35
& 30.00 $\pm$ 10.50
\\
    &7m8m & 3.12 $\pm$ 3.12 & \textbf{10.00} $\pm$ 8.67  &0.00 $\pm$ 0.00& 4.38 $\pm$ 3.56 & 8.75 $\pm$ 10.92
& \textbf{10.00} $\pm$ 13.51
\\ 
   & 8m9m & 6.88 $\pm$ 7.78 & 6.25 $\pm$ 7.65  &0.00 $\pm$ 0.00& 11.25 $\pm$ 13.18 & 13.12 $\pm$ 6.01
& \textbf{15.00} $\pm$ 18.01
\\   
   & 9m10m $^\diamond$& 64.38 $\pm$ 14.76 & 64.38 $\pm$ 12.62  &0.00 $\pm$ 0.00& 68.75 $\pm$ 15.93 & \textbf{74.38} $\pm$ 10.22
& 58.75 $\pm$ 16.15
\\
    &10m11m & 41.88 $\pm$ 11.18 & 43.12 $\pm$ 16.30  &0.00 $\pm$ 0.00& 21.25 $\pm$ 9.73 & 38.12 $\pm$ 20.66
& \textbf{50.62} $\pm$ 23.11
\\ 
   & 10m12m & 0.62 $\pm$ 1.40 & 0.62 $\pm$ 1.40  &0.00 $\pm$ 0.00& 0.00 $\pm$ 0.00 & \textbf{1.25} $\pm$ 2.80
& 0.00 $\pm$ 0.00
\\ 
   & 13m15m & 1.88 $\pm$ 1.71 & \textbf{0.62} $\pm$ 1.40  &0.00 $\pm$ 0.00& 1.25 $\pm$ 2.80 & 0.00 $\pm$ 0.00
& \textbf{0.62} $\pm$ 1.40
\\ 
    \midrule
 \multicolumn{2}{c}{\textbf{Average}}  & 42.08 $\pm$ 2.63 & 41.51 $\pm$ 3.53  &9.17 $\pm$ 1.78& 32.55 $\pm$ 4.31 & 47.60 $\pm$ 2.66& \textbf{48.12} $\pm$ 5.66\\
    \midrule
    \multirow{12}{*}{\rotatebox{90}{\makecell{Medium-Replay}}} 
    &3m $^\diamond$& 75.62 $\pm$ 7.13 & 80.00 $\pm$ 17.06  &81.88 $\pm$ 6.77& 75.00 $\pm$ 13.62 & 84.38 $\pm$ 8.56
& \textbf{85.00} $\pm$ 10.22
\\
    &4m & 83.12 $\pm$ 10.03 & 83.12 $\pm$ 8.44  &57.5 $\pm$ 25.64& 55.62 $\pm$ 34.40 & 75.00 $\pm$ 8.27
& \textbf{91.88} $\pm$ 9.78
\\
    &5m & 95.62 $\pm$ 5.23 & 95.62 $\pm$ 9.78  &89.38 $\pm$ 11.61& 93.12 $\pm$ 8.39 & 83.75 $\pm$ 13.51
& \textbf{100.00} $\pm$ 0.00
\\
    &10m & 87.50 $\pm$ 11.27 & 90.62 $\pm$ 11.05  &3.12 $\pm$ 6.99& 89.38 $\pm$ 8.44 & \textbf{93.12} $\pm$ 10.22
& 86.88 $\pm$ 7.78
\\
    &12m & 85.62 $\pm$ 10.96 & 88.12 $\pm$ 4.07  &1.25 $\pm$ 2.8& 70.62 $\pm$ 16.33 & 88.12 $\pm$ 10.22
& \textbf{94.38} $\pm$ 4.07
\\
    &5m6m $^\diamond$& 30.00 $\pm$ 13.55 & \textbf{33.12} $\pm$ 6.48  &18.75 $\pm$ 9.63& 9.38 $\pm$ 6.25 & 29.38 $\pm$ 13.37
& 13.75 $\pm$ 5.23
\\
    &7m8m & \textbf{14.38} $\pm$ 2.80 & 3.75 $\pm$ 2.61  &6.25 $\pm$ 4.94& 5.00 $\pm$ 7.84 & 9.38 $\pm$ 7.65
& 8.12 $\pm$ 7.19
\\ 
    &8m9m & 8.12 $\pm$ 3.56 & 11.88 $\pm$ 5.59  &1.25 $\pm$ 1.71& 1.88 $\pm$ 1.71 & 12.50 $\pm$ 4.94
& \textbf{26.88} $\pm$ 10.96
\\   
    &9m10m $^\diamond$ & 36.25 $\pm$ 14.08 & 42.50 $\pm$ 14.76  &0.62 $\pm$ 1.40& 14.38 $\pm$ 13.18 & \textbf{50.62} $\pm$ 13.51
& 43.75 $\pm$ 12.3
\\
    &10m11m  & 36.25 $\pm$ 8.15 & 29.38 $\pm$ 18.57  &0.00 $\pm$ 0.00& 32.50 $\pm$ 22.27 & \textbf{40.00} $\pm$ 13.69
& 38.12 $\pm$ 17.17
\\ 
    &10m12m & \textbf{2.50} $\pm$ 2.61 & 0.62 $\pm$ 1.40  &0.00 $\pm$ 0.00& 0.00 $\pm$ 0.00 & 0.62 $\pm$ 1.40
& 1.25 $\pm$ 2.80
\\ 
    &13m15m & 5.62 $\pm$ 2.61 & 5.00 $\pm$ 6.48  &0.00 $\pm$ 0.00& 0.62 $\pm$ 1.40 & 3.75 $\pm$ 4.07
& \textbf{8.12} $\pm$ 2.80
\\ 
    \midrule
   \multicolumn{2}{c}{\textbf{Average}} & 46.72 $\pm$ 2.35 & 46.98 $\pm$ 1.95  &21.67 $\pm$ 2.83& 37.29 $\pm$ 5.48 & 47.55 $\pm$ 3.56& \textbf{49.84} $\pm$ 3.36\\
    \midrule
    \multirow{12}{*}{\rotatebox{90}{\makecell{Medium-Expert}}} 
    &3m $^\diamond$& 74.38 $\pm$ 14.72 & 85.00 $\pm$ 19.44  &\textbf{100.00} $\pm$ 0.00& 53.75 $\pm$ 36.94 & 90.62 $\pm$ 19.26
& 98.75 $\pm$ 1.71
\\
    &4m & \textbf{95.00} $\pm$ 3.56 & 85.00 $\pm$ 23.11  &57.5 $\pm$ 29.86& 34.38 $\pm$ 45.93 & 94.38 $\pm$ 4.07
& 92.50 $\pm$ 2.80
\\
    &5m & 91.88 $\pm$ 9.78 & 58.12 $\pm$ 28.00  &80.62 $\pm$ 13.15& 31.25 $\pm$ 38.84 & \textbf{98.75} $\pm$ 2.80
& 93.12 $\pm$ 10.92
\\
    &10m & 93.75 $\pm$ 10.83 & 80.00 $\pm$ 19.71  &0.00 $\pm$ 0.00& 26.25 $\pm$ 34.13 & 95.62 $\pm$ 1.71
& \textbf{95.00} $\pm$ 5.68
\\
    &12m & 59.38 $\pm$ 43.24 & 73.12 $\pm$ 26.85  &0.00 $\pm$ 0.00& 3.75 $\pm$ 5.59 & 86.25 $\pm$ 13.73
& \textbf{90.00} $\pm$ 6.77
\\
    &5m6m $^\diamond$ & \textbf{43.12} $\pm$ 22.14 & 37.50 $\pm$ 18.62  &5.62 $\pm$ 5.59& 8.75 $\pm$ 8.09 & 42.50 $\pm$ 23.03
& 29.38 $\pm$ 17.06
\\
    &7m8m & 18.12 $\pm$ 4.64 & 16.25 $\pm$ 14.72  &0.00 $\pm$ 0.00& 4.38 $\pm$ 6.85 & \textbf{21.25} $\pm$ 8.67
& 19.38 $\pm$ 21.01
\\ 
    &8m9m & 31.25 $\pm$ 19.39 & 26.25 $\pm$ 18.96  &0.00 $\pm$ 0.00& 15.00 $\pm$ 21.58 & 28.12 $\pm$ 11.05
& \textbf{39.38} $\pm$ 26.57
\\   
    &9m10m $^\diamond$& 62.50 $\pm$ 24.90 & \textbf{84.38} $\pm$ 9.38  &0.00 $\pm$ 0.00& 23.75 $\pm$ 35.81 & 36.25 $\pm$ 20.20
& 40.62 $\pm$ 21.42
\\
    &10m11m& 51.88 $\pm$ 17.34 & 41.88 $\pm$ 19.21  &0.00 $\pm$ 0.00& 33.75 $\pm$ 38.87 & 42.50 $\pm$ 4.74
& \textbf{61.88} $\pm$ 20.18
\\ 
    &10m12m & 1.88 $\pm$ 4.19 & 2.50 $\pm$ 4.07  &0.00 $\pm$ 0.00& 0.00 $\pm$ 0.00 & 2.50 $\pm$ 5.59
& \textbf{4.38} $\pm$ 9.78
\\ 
    &13m15m & 0.00 $\pm$ 0.00 & 1.25 $\pm$ 1.71  &0.00 $\pm$ 0.00& 0.00 $\pm$ 0.00 & 0.62 $\pm$ 1.40
& \textbf{10.00} $\pm$ 10.69
\\ 
    \midrule
\multicolumn{2}{c}{\textbf{Average}}& 51.93 $\pm$ 7.21 & 49.27 $\pm$ 5.16  &20.31 $\pm$ 1.97& 19.58 $\pm$ 17.63 & 53.28 $\pm$ 1.88& \textbf{56.20} $\pm$ 3.21\\
    \bottomrule
    \multicolumn{7}{l}{\small $\diamond$ denotes the source tasks.} &\\
    \multicolumn{7}{l}{\small ``$X$m$Y$m'' represents ``$X$m\_vs\_$Y$m''.} &
    \end{tabular}
}
\end{center}
\vskip -0.1in
\end{table}

\begin{table}[htbp!]
\caption{Win rates of \textit{Stalker-Zealot} {scenario}. The reported results are averaged over 5 random seeds. The bold number denotes the best performance, and $\pm$ denotes one standard deviation.}
\label{tab:sz-win-rates}
\vskip 0.1in
\begin{center}
\resizebox{.9\linewidth}{!}{%
\begin{tabular}{clcccccc}
    \toprule
    \multicolumn{2}{c}{\textbf{Task}} & \textbf{BC-t} & \textbf{BC-r}  &\textbf{CQL}&\textbf{ODIS} & \textbf{HiSSD}& \textbf{SD-CQL (Ours)} \\
    \midrule
    \multirow{13}{*}{\rotatebox{90}{\makecell{Expert}}}
    &1s3z & 51.88 $\pm$ 28.26 & 65.00 $\pm$ 7.46  &70.00 $\pm$ 39.57& 41.25 $\pm$ 38.55 & 65.62 $\pm$ 19.76
& \textbf{74.38} $\pm$ 31.59
\\ 
    &1s4z & 34.38 $\pm$ 10.83 & \textbf{55.00} $\pm$ 31.14  &25.62 $\pm$ 20.30& 19.38 $\pm$ 29.18 & 52.50 $\pm$ 37.33
& 53.75 $\pm$ 28.07
\\ 
    &1s5z & 24.38 $\pm$ 28.42 & 19.38 $\pm$ 25.81  &1.25 $\pm$ 1.71& 30.62 $\pm$ 40.41 & 18.75 $\pm$ 22.10
& \textbf{60.00} $\pm$ 24.35
\\ 
    &2s3z $^\diamond$& \textbf{95.62} $\pm$ 2.80 & 95.00 $\pm$ 4.74  &36.88 $\pm$ 25.33& 77.50 $\pm$ 25.14 & 94.38 $\pm$ 7.78
& 85.62 $\pm$ 10.50
\\
    &2s4z $^\diamond$& \textbf{81.25} $\pm$ 7.33 & 76.88 $\pm$ 8.44  &1.25 $\pm$ 1.71& 53.12 $\pm$ 24.61 & 73.75 $\pm$ 5.68
& 71.88 $\pm$ 13.80
\\
    &2s5z & \textbf{66.25} $\pm$ 23.43 & 65.00 $\pm$ 18.54  &0.00 $\pm$ 0.00& 48.12 $\pm$ 20.44 & 61.88 $\pm$ 17.73
& 63.75 $\pm$ 10.50
\\ 
    &3s3z & \textbf{91.25} $\pm$ 5.59 & 76.88 $\pm$ 19.47  &3.12 $\pm$ 5.41& 81.88 $\pm$ 15.53 & 72.50 $\pm$ 23.53
& 90.00 $\pm$ 8.39
\\ 
    &3s4z & 93.12 $\pm$ 4.64 & 78.75 $\pm$ 6.01  &0.00 $\pm$ 0.00& 81.88 $\pm$ 8.95 & 93.12 $\pm$ 4.07
& \textbf{95.62} $\pm$ 5.23
\\ 
    &3s5z $^\diamond$& 93.75 $\pm$ 5.85 & 90.00 $\pm$ 8.39  &0.00 $\pm$ 0.00& 86.88 $\pm$ 5.13 & \textbf{95.00} $\pm$ 4.74
& 85.00 $\pm$ 11.14
\\ 
    &4s3z & 77.50 $\pm$ 17.31 & 78.12 $\pm$ 16.97  &0.00 $\pm$ 0.00& 71.25 $\pm$ 19.06 & 83.75 $\pm$ 15.05
& \textbf{90.00} $\pm$ 9.48
\\ 
    &4s4z & 66.25 $\pm$ 16.89 & 52.50 $\pm$ 8.39  &0.00 $\pm$ 0.00& 52.50 $\pm$ 20.89 & 60.62 $\pm$ 34.28
& \textbf{74.38} $\pm$ 13.51
\\ 
    &4s5z & 54.38 $\pm$ 12.81 & 38.75 $\pm$ 3.56  &0.00 $\pm$ 0.00& 35.00 $\pm$ 27.99 & 52.50 $\pm$ 22.90
& \textbf{71.88} $\pm$ 17.95
\\ 
    &4s6z & 61.88 $\pm$ 10.92 & 34.38 $\pm$ 13.62  &0.00 $\pm$ 0.00& 36.25 $\pm$ 21.83 & 41.25 $\pm$ 21.36
& \textbf{68.12} $\pm$ 14.89
\\
    \midrule
\multicolumn{2}{c}{\textbf{Average}} & 68.61 $\pm$ 3.93 & 63.51 $\pm$ 2.64  &10.62 $\pm$ 5.88& 55.05 $\pm$ 9.11 & 66.59 $\pm$ 7.77& \textbf{75.72} $\pm$ 3.73\\
    \midrule
    \multirow{13}{*}{\rotatebox{90}{\makecell{Medium}}} 
    &1s3z & 3.12 $\pm$ 3.83 & 14.38 $\pm$ 25.73  &36.25 $\pm$ 25.25& 10.62 $\pm$ 22.05 & 35.0 $\pm$ 22.69
& \textbf{64.38} $\pm$ 29.70
\\ 
    &1s4z & 18.75 $\pm$ 25.86 & 11.88 $\pm$ 24.84  &43.75 $\pm$ 29.89& 7.50 $\pm$ 8.15 & 13.75 $\pm$ 5.23
& \textbf{48.75} $\pm$ 37.80
\\ 
    &1s5z & 5.62 $\pm$ 6.77 & 3.75 $\pm$ 5.13  &11.88 $\pm$ 13.33& 0.62 $\pm$ 1.40 & 9.38 $\pm$ 6.25
& \textbf{51.25} $\pm$ 38.76
\\ 
    &2s3z $^\diamond$& 49.38 $\pm$ 13.51 & 48.12 $\pm$ 4.74  &\textbf{62.5} $\pm$ 14.32& 38.12 $\pm$ 11.57 & 50.62 $\pm$ 10.22
& 30.63 $\pm$ 23.94
\\
    &2s4z $^\diamond$ & 11.88 $\pm$ 8.09 & 8.75 $\pm$ 13.15  &23.12 $\pm$ 16.03& 8.75 $\pm$ 7.78 & 9.38 $\pm$ 7.65
& \textbf{29.38} $\pm$ 16.92
\\
    &2s5z & 13.75 $\pm$ 10.50 & 16.25 $\pm$ 8.67  &1.25 $\pm$ 2.8& 18.75 $\pm$ 27.15 & 14.37 $\pm$ 6.48
& \textbf{42.50} $\pm$ 23.24
\\ 
    &3s3z & 39.38 $\pm$ 15.56 & 26.25 $\pm$ 15.56  &16.25 $\pm$ 22.25& 21.88 $\pm$ 20.01 & 31.87 $\pm$ 16.30
& \textbf{45.00} $\pm$ 12.62
\\ 
    &3s4z & 46.25 $\pm$ 17.59 & 23.75 $\pm$ 12.22  &3.75 $\pm$ 8.39& 18.12 $\pm$ 17.59 & 45.62 $\pm$ 14.59
& \textbf{64.38} $\pm$ 28.35
\\ 
    &3s5z $^\diamond$& 41.88 $\pm$ 9.78 & 43.12 $\pm$ 18.14  &0.00 $\pm$ 0.00& 9.38 $\pm$ 9.38 & 42.50 $\pm$ 16.62
& \textbf{56.25} $\pm$ 9.11
\\ 
    &4s3z & 37.50 $\pm$ 26.42 & 42.50 $\pm$ 18.57  &0.00 $\pm$ 0.00& 11.88 $\pm$ 8.67 & 33.12 $\pm$ 13.91
& \textbf{48.12} $\pm$ 16.92
\\ 
    &4s4z & \textbf{23.75} $\pm$ 5.23 & 16.25 $\pm$ 14.56  &0.00 $\pm$ 0.00& 8.12 $\pm$ 10.03 & 7.50 $\pm$ 3.56
& 18.12 $\pm$ 10.69
\\ 
    &4s5z & \textbf{10.62} $\pm$ 5.68 & 10.00 $\pm$ 9.73  &0.00 $\pm$ 0.00& 0.62 $\pm$ 1.40 & 4.38 $\pm$ 4.19
& \textbf{10.62} $\pm$ 7.84
\\ 
    &4s6z & 9.38 $\pm$ 4.94 & 5.62 $\pm$ 4.64  &0.00 $\pm$ 0.00& 1.25 $\pm$ 2.80 & 8.12 $\pm$ 4.74
& \textbf{16.25} $\pm$ 11.35
\\
    \midrule
 \multicolumn{2}{c}{\textbf{Average}}  & 23.94 $\pm$ 2.72 & 20.82 $\pm$ 3.78  &15.29 $\pm$ 4.89& 11.97 $\pm$ 7.53 & 23.51 $\pm$ 3.76& \textbf{40.43} $\pm$ 7.06\\
    \midrule
    \multirow{13}{*}{\rotatebox{90}{\makecell{Medium-Replay}}} 
    &1s3z & 17.50 $\pm$ 11.40 & 33.12 $\pm$ 20.32  &18.75 $\pm$ 14.66& 3.12 $\pm$ 5.41 & \textbf{55.62} $\pm$ 26.46
& 21.25 $\pm$ 14.39
\\ 
    &1s4z & 7.50 $\pm$ 6.09 & 13.75 $\pm$ 13.00  &32.5 $\pm$ 16.18& 10.62 $\pm$ 10.73 & \textbf{15.00} $\pm$ 11.98
& 13.75 $\pm$ 24.06
\\ 
    &1s5z & 3.75 $\pm$ 4.07 & 3.75 $\pm$ 8.39  &\textbf{10.00} $\pm$ 9.48& 5.00 $\pm$ 7.19 & \textbf{10.00} $\pm$ 13.87
& 8.75 $\pm$ 8.39
\\ 
    &2s3z $^\diamond$& 12.50 $\pm$ 6.63 & 6.88 $\pm$ 5.59  &\textbf{53.75} $\pm$ 13.51& 12.50 $\pm$ 16.39 & 12.50 $\pm$ 6.99
& 7.50 $\pm$ 2.80
\\
    &2s4z $^\diamond$& 8.12 $\pm$ 9.53 & 10.62 $\pm$ 5.68  &\textbf{18.12} $\pm$ 11.98& 12.50 $\pm$ 13.44 & 5.62 $\pm$ 4.07
& 6.25 $\pm$ 4.42
\\
    &2s5z & 11.25 $\pm$ 9.27 & \textbf{22.50} $\pm$ 24.25  &9.38 $\pm$ 8.27& 10.62 $\pm$ 14.59 & 9.38 $\pm$ 6.25
& 17.50 $\pm$ 15.08
\\ 
    &3s3z & 13.75 $\pm$ 12.62 & 12.50 $\pm$ 4.94  &22.50 $\pm$ 7.13& 12.50 $\pm$ 13.44 & 18.75 $\pm$ 24.11
& \textbf{57.50} $\pm$ 15.87
\\ 
    &3s4z & 27.50 $\pm$ 17.46 & 45.00 $\pm$ 19.09  &13.12 $\pm$ 11.57& 8.12 $\pm$ 9.00 & 20.62 $\pm$ 10.50
& \textbf{47.50} $\pm$ 32.43
\\ 
    &3s5z $^\diamond$&26.88 $\pm$ 13.91 & 30.00 $\pm$ 11.18  &3.75 $\pm$ 5.13& 5.62 $\pm$ 3.42 & 13.75 $\pm$ 7.84
& \textbf{40.62} $\pm$ 8.27
\\ 
    &4s3z & 11.25 $\pm$ 18.57 & \textbf{23.75} $\pm$ 19.59  &4.38 $\pm$ 5.23& 13.75 $\pm$ 16.18 & 15.00 $\pm$ 18.67
& \textbf{23.75} $\pm$ 21.49
\\ 
    &4s4z & 12.50 $\pm$ 9.11 & 18.75 $\pm$ 15.15  &2.50 $\pm$ 3.42& 3.75 $\pm$ 3.42 & 11.88 $\pm$ 8.09
& \textbf{28.12} $\pm$ 15.93
\\ 
    &4s5z & 6.25 $\pm$ 3.83 & 5.00 $\pm$ 6.48  &0.00 $\pm$ 0.00& 5.00 $\pm$ 8.15 & 1.88 $\pm$ 2.80
& \textbf{16.88} $\pm$ 9.00
\\ 
    &4s6z & 3.75 $\pm$ 5.13 & 3.75 $\pm$ 4.07  &0.62 $\pm$ 1.40& 3.12 $\pm$ 3.83 & 5.00 $\pm$ 1.71
& \textbf{10.62} $\pm$ 6.09
\\
    \midrule
   \multicolumn{2}{c}{\textbf{Average}} & 12.50 $\pm$ 1.75 & 17.64 $\pm$ 4.40  &14.57 $\pm$ 2.02& 8.17 $\pm$ 6.08 & 15.00 $\pm$ 4.02& \textbf{23.08} $\pm$ 3.38\\
    \midrule
    \multirow{13}{*}{\rotatebox{90}{\makecell{Medium-Expert}}} 
    &1s3z & 47.50 $\pm$ 39.12 & 25.00 $\pm$ 30.22  &\textbf{80.00} $\pm$ 20.20& 49.38 $\pm$ 36.74 & 58.75 $\pm$ 35.93
& 73.75 $\pm$ 40.42
\\ 
    &1s4z & 5.00 $\pm$ 3.56 & 13.12 $\pm$ 9.73  &43.12 $\pm$ 18.67& 6.88 $\pm$ 15.37 & 10.00 $\pm$ 6.01
& \textbf{72.50} $\pm$ 27.72
\\ 
    &1s5z & 11.88 $\pm$ 23.22 & 10.62 $\pm$ 20.32  &1.88 $\pm$ 1.71& 0.62 $\pm$ 1.40 & 6.25 $\pm$ 12.30
& \textbf{72.50} $\pm$ 17.73
\\ 
    &2s3z $^\diamond$& 80.62 $\pm$ 21.47 & 71.25 $\pm$ 10.22  &49.38 $\pm$ 11.35& 59.38 $\pm$ 35.63 & 68.75 $\pm$ 14.99
& \textbf{86.25} $\pm$ 10.50
\\
    &2s4z $^\diamond$& 37.50 $\pm$ 32.10 & 61.25 $\pm$ 15.87  &0.00 $\pm$ 0.00& 47.50 $\pm$ 31.44 & 56.25 $\pm$ 24.71
& \textbf{62.50} $\pm$ 19.64
\\
    &2s5z & 25.62 $\pm$ 19.94 & 21.88 $\pm$ 11.48  &0.00 $\pm$ 0.00& 27.50 $\pm$ 18.14 & 23.12 $\pm$ 9.53
& \textbf{40.62} $\pm$ 13.80
\\ 
    &3s3z & 47.50 $\pm$ 37.85 & 56.88 $\pm$ 11.57  &0.00 $\pm$ 0.00& 55.00 $\pm$ 29.20 & 63.75 $\pm$ 29.37
& \textbf{71.25} $\pm$ 16.00
\\ 
    &3s4z & 60.00 $\pm$ 17.87 & 74.38 $\pm$ 12.77  &0.00 $\pm$ 0.00& 50.00 $\pm$ 30.78 & 72.50 $\pm$ 16.15
& \textbf{93.12} $\pm$ 3.42
\\ 
    &3s5z $^\diamond$&67.50 $\pm$ 17.76 & 61.88 $\pm$ 12.18  &0.00 $\pm$ 0.00& 31.88 $\pm$ 24.45 & 50.62 $\pm$ 18.01
& \textbf{87.50} $\pm$ 6.99
\\ 
    &4s3z & 62.50 $\pm$ 18.62 & \textbf{70.62} $\pm$ 7.84  &0.00 $\pm$ 0.00& 39.38 $\pm$ 27.12 & 55.00 $\pm$ 24.27
& 55.00 $\pm$ 28.86
\\ 
    &4s4z & 30.62 $\pm$ 5.59 & 35.00 $\pm$ 23.11  &0.00 $\pm$ 0.00& 19.38 $\pm$ 12.77 & 41.25 $\pm$ 21.36
& \textbf{61.88} $\pm$ 20.54
\\ 
    &4s5z & 10.00 $\pm$ 5.13 & 6.88 $\pm$ 10.22  &0.00 $\pm$ 0.00& 6.88 $\pm$ 2.61 & 7.50 $\pm$ 6.48
& \textbf{49.38} $\pm$ 26.28
\\ 
    &4s6z & 10.00 $\pm$ 7.46 & 6.25 $\pm$ 5.85  &0.00 $\pm$ 0.00& 5.62 $\pm$ 7.78 & 10.00 $\pm$ 5.13
& \textbf{44.38} $\pm$ 19.94
\\
    \midrule
\multicolumn{2}{c}{\textbf{Average}}& 38.17 $\pm$ 5.38 & 39.62 $\pm$ 2.98  &13.41 $\pm$ 2.53& 30.72 $\pm$ 14.65 & 40.29 $\pm$ 4.64& \textbf{66.97} $\pm$ 7.01\\
    \bottomrule
    \multicolumn{7}{l}{\small $\diamond$ denotes the source tasks.} &
    \end{tabular}
}
\end{center}
\vskip -0.5in
\end{table}

\subsection{One-to-Multi Task Sets}\label{app:results-otm}
We present the detailed results of our evaluation experiments on the One-to-Multi task sets in Table \ref{tab:ssi-win-rates}. The source tasks used for training are marked by ``\(\diamond\)''. The results show that, despite being trained on only one source task, SD-CQL exhibits better multi-task generalization, as reflected in its exceptional average performance across all test tasks.
\begin{table*}[!ht]
\caption{Win rates of One-to-Multi task sets. The reported results are averaged over 5 random seeds. The bold number denotes the best performance, and $\pm$ denotes one standard deviation.}\label{tab:ssi-win-rates}
\begin{center}
\resizebox{\linewidth}{!}{%
\begin{tabular}{clccccccc}
\toprule
\multicolumn{2}{c}{\textbf{Task Set}} & \textbf{BC-t} & \textbf{BC-r} & \textbf{CQL} &\textbf{ODIS} & \textbf{HiSSD} & \textbf{SD-CQL (Ours)} \\
\midrule
\multirow{5}{*}{\rotatebox{90}{\makecell{\textit{Marine}\\\textit{Single}}}} 
    &3m $^\diamond$ & \textbf{100.00} $\pm$ 0.00 & 96.88 $\pm$ 4.42 &\textbf{100.00} $\pm$ 0.00& 99.38 $\pm$ 1.4 & \textbf{100.00} $\pm$ 0.00 & \textbf{100.00} $\pm$ 0.00\\
    &5m & 81.88 $\pm$ 9.22 & 67.5 $\pm$ 27.3 & 88.12 $\pm$ 7.46 &37.5 $\pm$ 33.15 & 66.88 $\pm$ 32.52 & \textbf{91.25} $\pm$ 10.92\\
    &8m & 38.75 $\pm$ 22.38 & 35.62 $\pm$ 29.37 &40.00 $\pm$ 26.00 &11.88 $\pm$ 21.47 & 31.87 $\pm$ 41.01 & \textbf{49.38} $\pm$ 23.63\\
    &10m & 20.62 $\pm$ 22.38 & 16.25 $\pm$ 20.54 & 34.38 $\pm$ 25.39  &11.25 $\pm$ 25.16 & 18.75 $\pm$ 40.20 & \textbf{41.88} $\pm$ 25.83\\
    &12m & 9.38 $\pm$ 13.98 & 2.50 $\pm$ 5.59 & 17.50 $\pm$ 10.27 &6.25 $\pm$ 13.98 & 3.12 $\pm$ 6.99 & \textbf{27.50} $\pm$ 21.13\\
    \midrule
    \multicolumn{2}{c}{\textbf{Average}} & 50.12 $\pm$ 9.65 & 43.75 $\pm$ 14.42 & 56.00 $\pm$ 11.07 &33.25 $\pm$ 16.50 & 44.12 $\pm$ 18.89 & \textbf{62.00} $\pm$ 13.80\\
\midrule\midrule
\multirow{5}{*}{\rotatebox{90}{\makecell{\textit{Marine} \\ \textit{Single-Inv}}}}
    &10m $^\diamond$ & \textbf{100.00} $\pm$ 0.00 & 98.75 $\pm$ 2.80 &0.62 $\pm$ 1.4 &   88.75 $\pm$ 12.22 & \textbf{100.00} $\pm$ 0.00 & \textbf{100.00} $\pm$ 0.00\\
    &8m &  98.75 $\pm$ 1.71 & 83.12 $\pm$ 29.28 & 3.75 $\pm$ 8.39 &90.62 $\pm$ 8.84 & \textbf{100.00} $\pm$ 0.00 & \textbf{100.00} $\pm$ 0.00\\
    &5m & 3.75 $\pm$ 6.77 & 33.12 $\pm$ 33.63 & 6.25 $\pm$ 13.98 &87.50 $\pm$ 8.84  & 91.88 $\pm$ 6.09 & \textbf{97.50} $\pm$ 2.61\\
    &4m & 1.25 $\pm$ 2.80 & 20.62 $\pm$ 33.04 & 0.62 $\pm$ 1.4 &62.50 $\pm$ 21.99 & 73.75 $\pm$ 15.72 & \textbf{75.62 }$\pm$ 21.58\\
    &3m & 0.00 $\pm$ 0.00  & 6.88 $\pm$ 13.69 & 2.5 $\pm$ 4.07 &12.50 $\pm$ 18.22 & \textbf{68.75} $\pm$ 17.4 & 64.38 $\pm$ 15.4\\
    \midrule
    \multicolumn{2}{c}{\textbf{Average}} & 40.75 $\pm$ 1.83 & 48.50 $\pm$ 17.09 & 2.75 $\pm$ 5.19 &68.38 $\pm$ 11.47 & 86.88 $\pm$ 4.49 & \textbf{87.50} $\pm$ 6.54\\
\bottomrule
\multicolumn{6}{l}{\small $\diamond$ denotes the source task.}\\
\end{tabular}
}
\end{center}
\end{table*}

\newpage

\section{Learning Curves}\label{app:curve}
\begin{figure}[!htbp!]
\vskip 0.2in
\begin{center}
\centerline{\includegraphics[width=\textwidth]{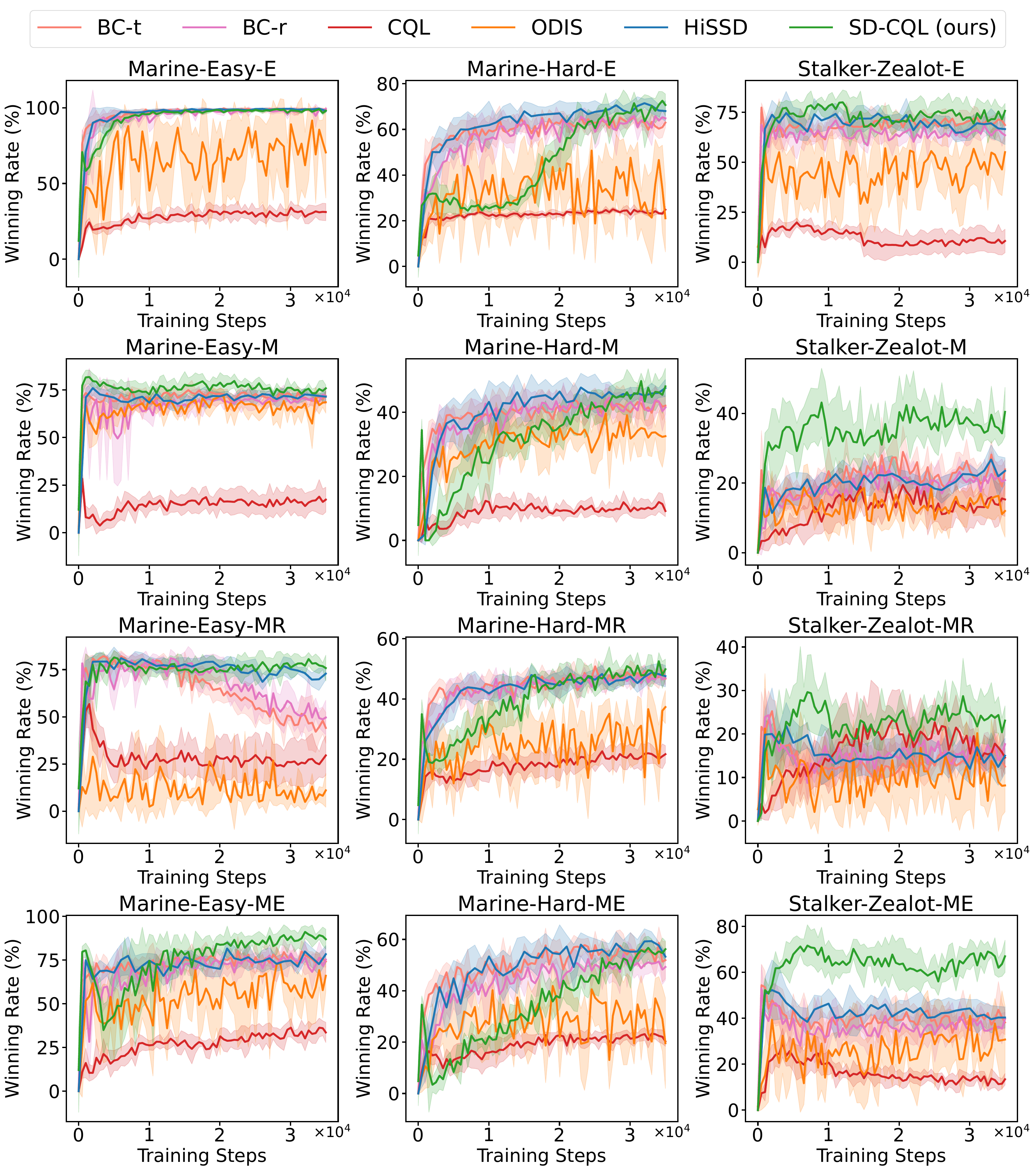}}
\caption{Average winning rates on Multi-to-Multi task sets. We report results over 5 random seeds, where the solid line represents the mean and the shaded area represents one standard deviation.}
\label{fig:mtm}
\end{center}
\vskip -0.2in
\end{figure}

In Figures \ref{fig:mtm}, we plot the learning curves of the average performance across different task sets for Multi-to-Multi. To make the figures clearer, we use abbreviations to represent the dataset quality. Specifically: -E, -M, -MR, and -ME stand for expert, medium, medium-replay, and medium-expert, respectively. We report results over 5 random seeds, where the solid line represents the mean and the shaded area represents one standard deviation. It is evident that SD-CQL sustains the highest average performance across most task sets.

\begin{figure}[!ht]
\vskip 0.2in
\begin{center}
\centerline{\includegraphics[width=\textwidth]{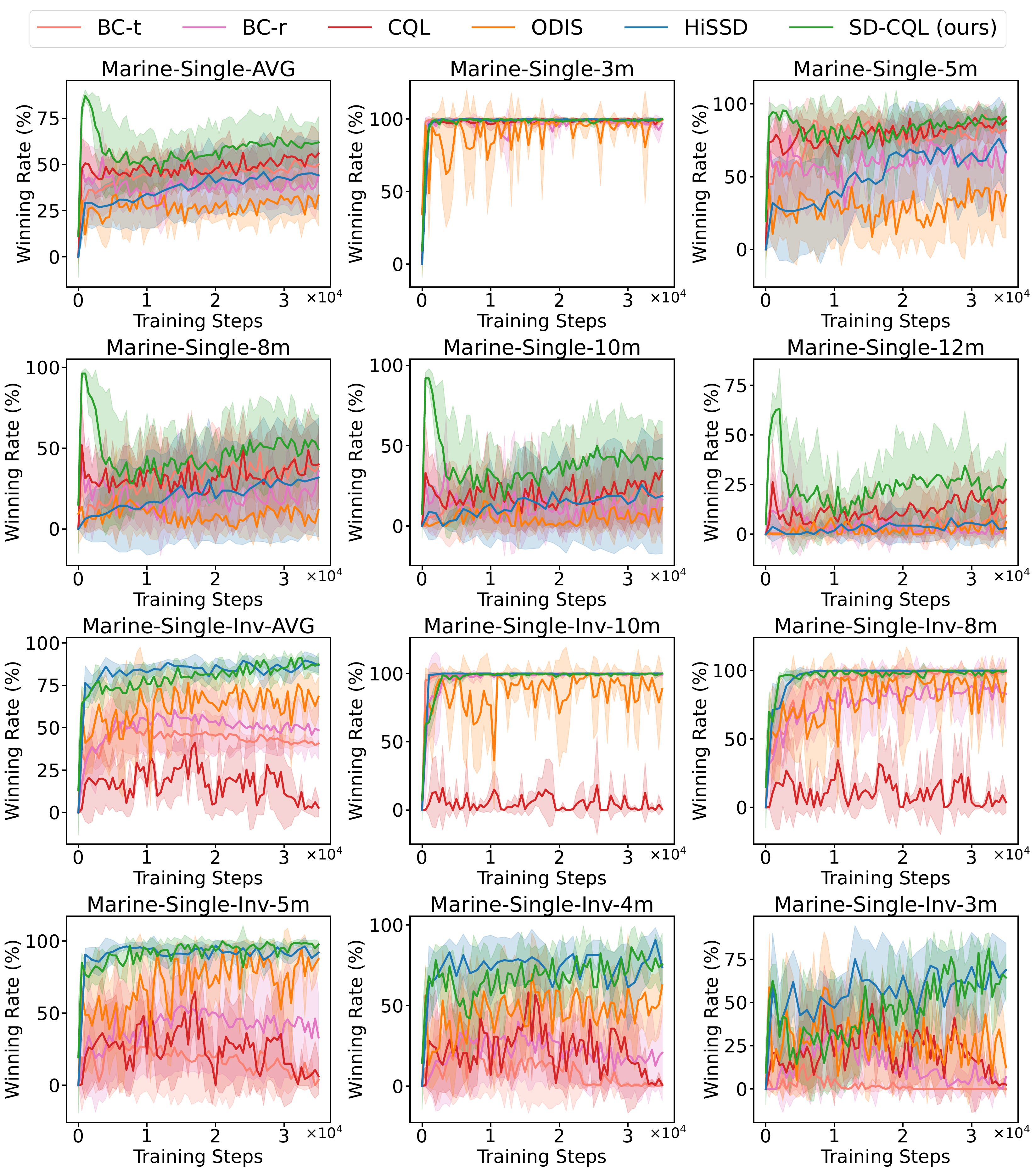}}
\caption{Average and task-specific winning rates on One-to-Multi task sets. We report results over 5 random seeds, where the solid line represents the mean and the shaded area represents one standard deviation.}
\label{fig:otm}
\end{center}
\vskip -0.2in
\end{figure}

\newpage
In Figure \ref{fig:otm}, we plot the learning curves for the average performance (marked as ``AVG'') and task-specific performance (marked with the respective task name) for One-to-Multi task sets. The results are reported over 5 random seeds, with the solid line representing the mean and the shaded area indicating one standard deviation. It can be observed that even when trained on a single source task, SD-CQL exhibits the best multi-task generalization performance. Notably, in the \textit{Marine-Single} task set, SD-CQL's peak performance on unseen tasks significantly surpasses that of the other baselines.

\end{document}